\documentclass[letterpaper, 10 pt, conference]{ieeeconf}
\IEEEoverridecommandlockouts 
\usepackage{cite}
\usepackage{amsmath,amssymb,amsfonts}
\usepackage{algorithmic}
\usepackage{graphicx}
\usepackage{textcomp}
\usepackage{xcolor}
\usepackage{subcaption}

\newcommand{\namePaper}{LADFN}
\newcommand{\nameLOAM}{LOAM}
\newcommand{\nameLIDAR}{LIDAR}
\newcommand{\nameCARLA}{CARLA}

\newcommand{\eqnRLSol}{
\begin{equation} \label{eq:RLSol}
\small
    V_\pi(s_0) = \mathbb{E}\bigg(    \sum_{t=0}^{\infty} \gamma^t \mathrm{R}  \Big(s_{t+1}, s_t, \pi\big(s_t\big)\Big)     \bigg),
\end{equation}
}

\newcommand{\eqnReward}{
\begin{equation} \label{eq:TotalReward}
    \small
    \mathcal{R} = \mathcal{P}\times\mathcal{T} + (1-\mathcal{P})\times\mathcal{F} + \mathcal{G},
\end{equation}
}

\newcommand{\eqnFeatureReward}{
\begin{equation} \label{eq:FeatReward}
\small
\begin{split}
    \mathcal{F} & = K_1 \times \bigg (y_c^O \times \Big (\frac{2\times(y_e-{y_e}_\mathrm{min})}{{y_e}_\mathrm{max} - {y_e}_\mathrm{min}} \Big) - 1)\bigg)\\
    & - K_2 \times (1- \vert y_c^O \vert)\times\vert a_y \vert,
\end{split}
\end{equation}
}

\newcommand{\eqnTrafficReward}{
\begin{equation} \label{eq:TrafficReward}
\small
\mathcal{T} = \sum_{i=1}^{n} p_i \Big[\big(K_3\times \vert a_x-v_{tr_i} \vert\big)+ K_4\times\big(\vert y_{tr_i}^{t} - y_e^t \vert - \vert y_{tr_i}^{t-1}-y_{e}^{t-1}\vert\big)\Big],
\end{equation} 
}

\newcommand{\figPath}{figures/General/PNG}
\newcommand{\figMainResultsPath}{figures/Quantitative/PNG}
\newcommand{\SubFigureScalar}{0.25\textwidth}
\newcommand{\GraphicsScalar}{1\linewidth}

\newcommand{\figTeaserCaption}{\small \textbf{Active navigation significantly reduces autonomous vehicle drift.} \textbf{Left column} represents baseline (Stanley Control) and \textbf{right column} represents our approach (\namePaper). \textbf{Top Panel:} CARLA \cite{Dosovitskiy17} environments showing ego vehicles planning their trajectories. \textbf{Bottom Panel:} LOAM \cite{zhang2014loam} map visualizations of the same scenes. The green lines delineate ground-truth trajectories, and the purple lines delineate \nameLOAM's generated odometry.  While drift from the green ground-truth is pronounced on the left, with \namePaper\ there is no such drift or deviation on the right. For a \textbf{100m} run, a Stanley-controlled ego drifts in the $xy$-plane by \textbf{14.05m}, eventually colliding with nearby traffic, whereas an ego controlled by our approach drifts by \textbf{0.62m}, due to traffic collision-avoidance and active navigation towards nearby features. This drift also affects \nameLOAM's \cite{zhang2014loam} mapping potential: the map on the left is distorted due to the ego's haphazard odometry, whereas the map on the right remains relatively unaffected.}

\newcommand{\figPipelineCaption}{\small \textbf{Pipeline of our approach}. First, a map of the input CARLA \cite{Dosovitskiy17} scene is built using LOAM \cite{zhang2014loam}. Filtering is applied to remove dynamic traffic objects, if present. The state obtained from this map is passed into a deep RL neural network that performs Proximal Policy Optimization \cite{schulman2017proximal}, to help choose actions $a \in \mathcal{A}$, our action space. This model ensures that the ego drives towards feature-rich lanes and avoids traffic in its way. The action inferences from the model are then are used to generate a set of discrete waypoints for the ego to traverse. A cubic spline is fit on these waypoints to generate a smooth trajectory, which is then tracked using a Stanley Controller \cite{stanleycontroller}.}

\newcommand{\figRLFrameCaption}{\small Our \textit{RL Frame} representation in a CARLA scene.}

\newcommand{\figFeatureProxCaption}{\small \textbf{Benefits of active navigation towards feature-rich regions.} More \nameLIDAR\ rays from $\mathbf{ego_1}$ than $\mathbf{ego_2}$ interact with the texture-rich region $T_R$, since $\theta_1 > \theta_2$, resulting in a more dense reconstruction.}

\newcommand{\figQuanResultsBarsOnlyCaption}{\small \textbf{\namePaper\ significantly reduces drift than competing benchmarks on a diversity of CARLA scenes.} Firstly, we want to show that \namePaper\ leads to lower average drift than the benchmark. This is clearly shown in the \textbf{first column} where \namePaper\ reduces Stanley's drift by at least $1.48\times$. Secondly, we provide a framework for analyzing the performance of goal-reaching tasks, seen in the final translation offset plots in the \textbf{second column}, where once again, \namePaper\ consistently outperforms the Vanilla Stanley controller by at least $1.63\times$. Thirdly, we show that rotational errors are significantly reduced by \namePaper\ in the \textbf{third column}. In dynamic scenes (2,3,5), dynamic obstacle filtering was enabled. In \textbf{all} columns, \textit{F} in the plot legends denotes that filtering of dynamic obstacles was carried out. Diagonal hatches '/' inside bars represent collision occurrence of the ego with its surroundings in the experiment.}

\newcommand{\figQualResultsCaption}{\small \textbf{\namePaper\ significantly reduces drift with a minor change in the path taken to reach the goal}. The \nameLOAM-generated trajectories are shown in red, the ground-truth trajectories are shown in black, and the green structures correspond to feature-rich regions. The left column shows \textit{Vanilla Stanley + F} while the right column shows \textit{\namePaper\ + F}. Each row corresponds to a different scene. These scenes were run with a \nameLIDAR\ range of $45$ meters and dynamic obstacle filtering enabled.}

\newcommand{\figTeaser}{
\begin{figure}[!htb]
\begin{subfigure}[b]{0.23\textwidth}
  \centering
  \includegraphics[width=\linewidth]{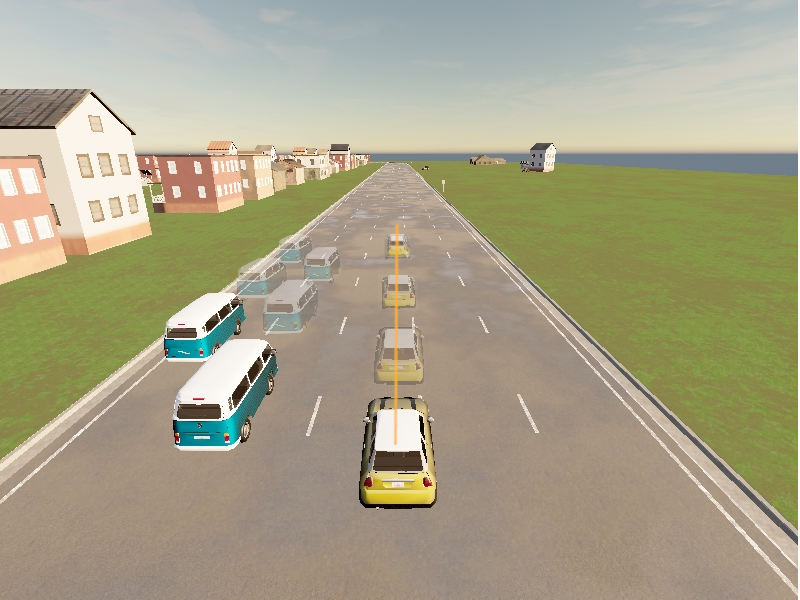}
\end{subfigure}
\hfill
\begin{subfigure}[b]{0.23\textwidth}
  \centering
  \includegraphics[width=\linewidth]{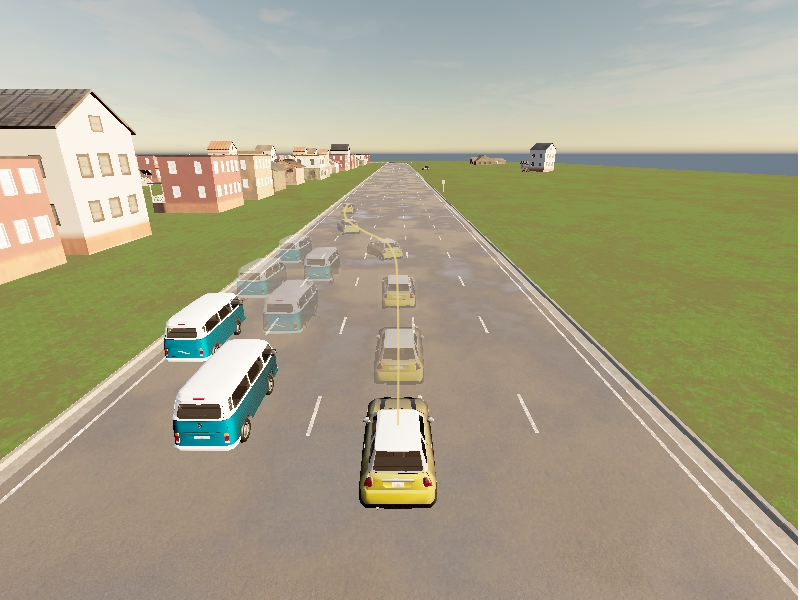}
\end{subfigure}
\newline
\begin{subfigure}[b]{0.23\textwidth}
  \centering
  \includegraphics[width=\linewidth]{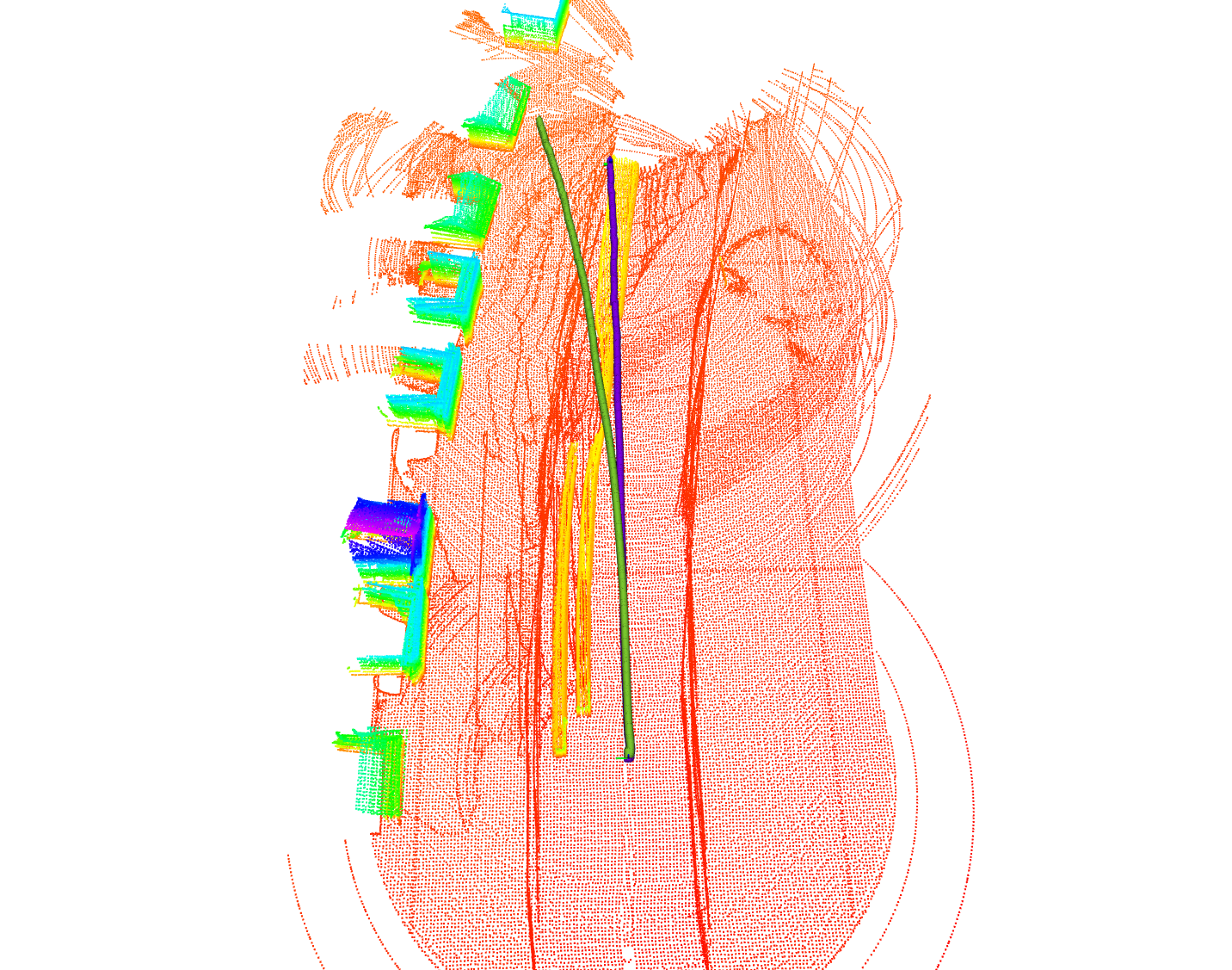}
\end{subfigure}
\hfill
\begin{subfigure}[b]{0.23\textwidth}
  \centering
  \includegraphics[width=\linewidth]{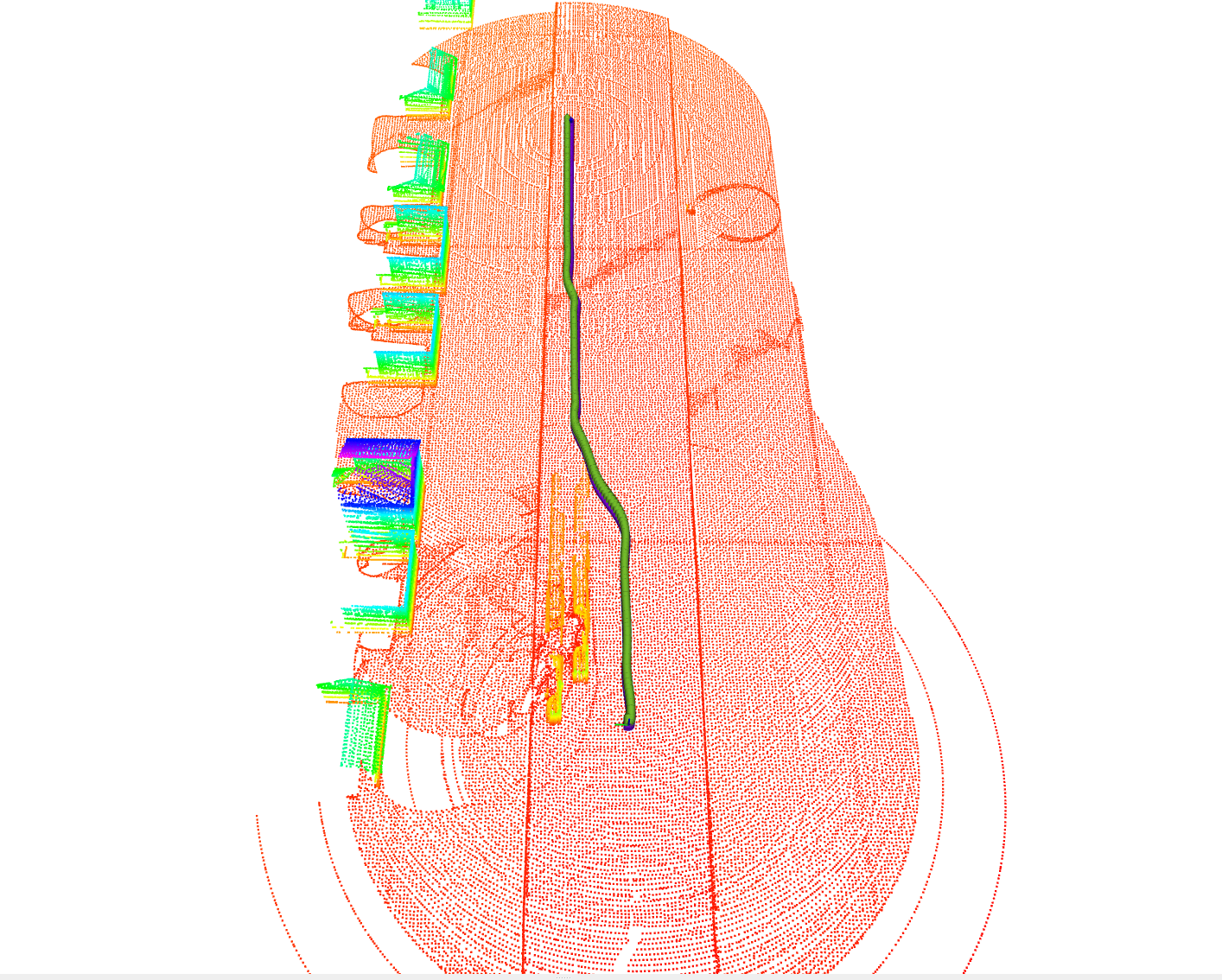}  
\end{subfigure}
\setlength{\belowcaptionskip}{-15pt}
\caption{\figTeaserCaption}
\label{fig:Teaser}
\end{figure}
}

\newcommand{\figPipeline}{
\begin{figure*}[!htb]
    \includegraphics[width=\textwidth, height=7cm]{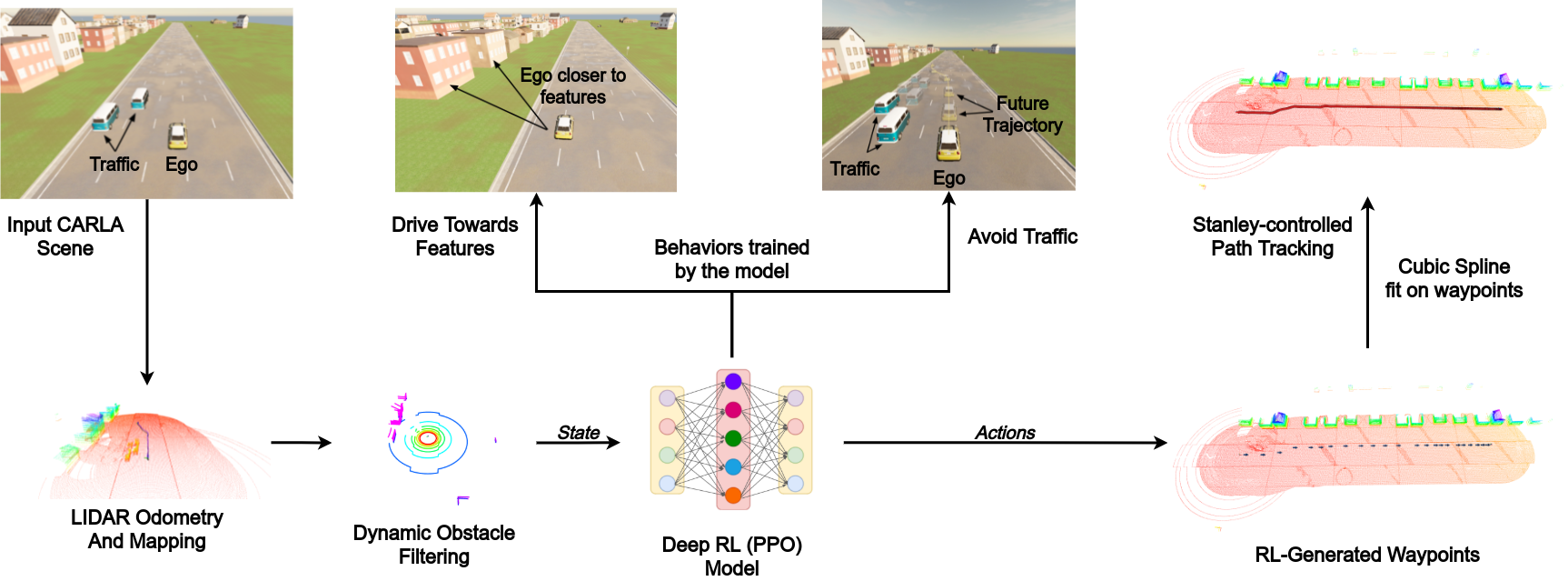}
    \centering
    \setlength{\belowcaptionskip}{-15pt}
    \caption{\figPipelineCaption}
    \label{fig:Pipeline}
\end{figure*}
}

\newcommand{\figRLFrame}{
\begin{figure}[h]
    \includegraphics[scale=0.65, width=\linewidth]{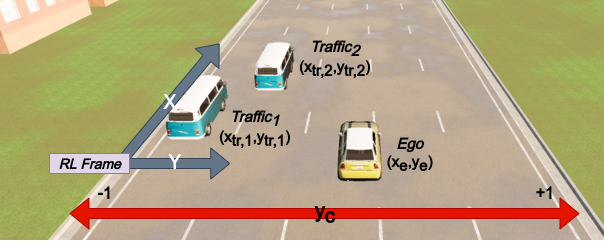}
    \centering
    \setlength{\abovecaptionskip}{-5pt}
    \setlength{\belowcaptionskip}{-15pt}
    \caption{\figRLFrameCaption}
    \label{fig:rlframe}
\end{figure}
}

\newcommand{\figFeatureProx}{
\begin{figure}[h]
    \includegraphics[height=4.5cm]{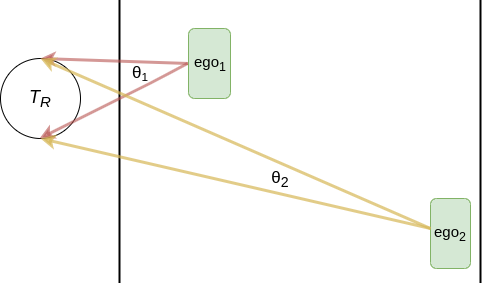}
    \centering
    \setlength{\belowcaptionskip}{-5pt}
    \caption{\figFeatureProxCaption}
    \label{fig:featprox}
\end{figure}
}

\newcommand{\figQualResults}{
\begin{figure}[htbp]
    \includegraphics[width=\linewidth, height=0.85\textheight]{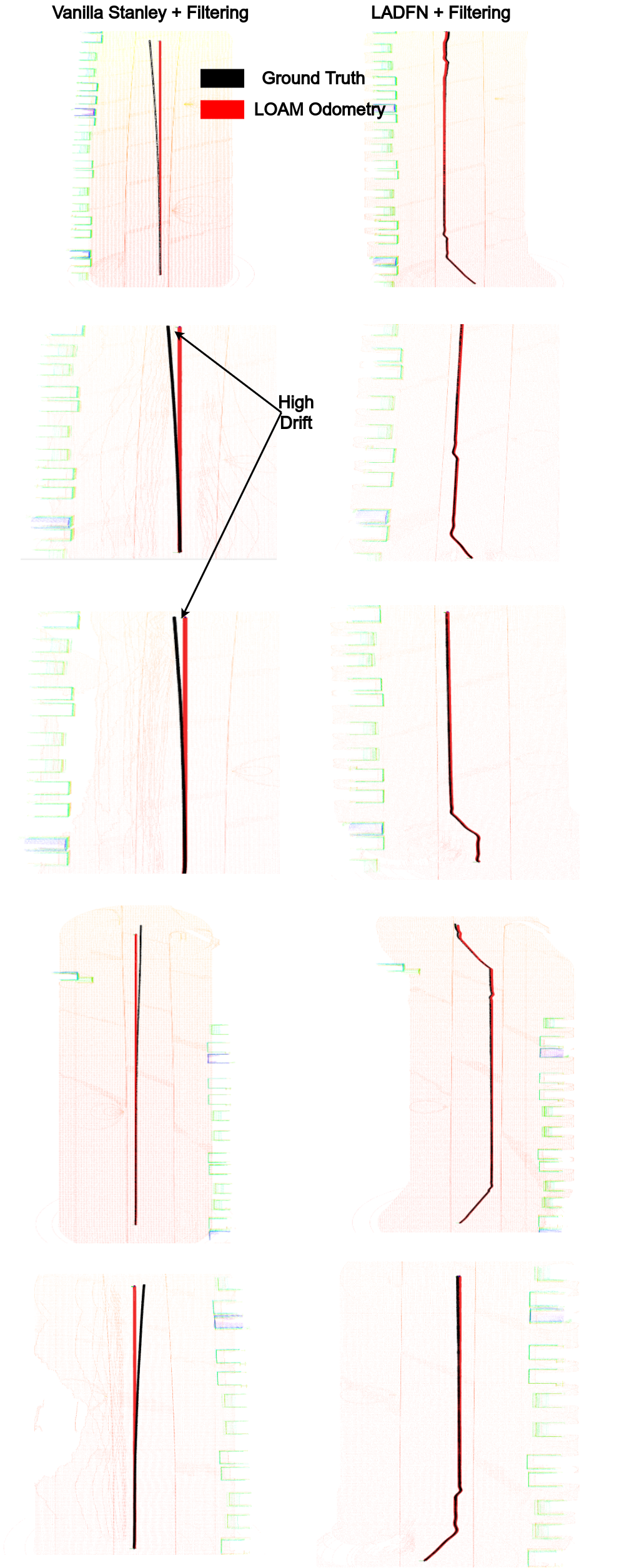}
\setlength{\belowcaptionskip}{-18pt}
\caption{\figQualResultsCaption}
\label{fig:QualResults}
\end{figure}
}

\newcommand{\figQuanResultsBarsOnly}{
\begin{figure*}[!htb]
\begin{subfigure}[b]{\SubFigureScalar}
  \centering
  \includegraphics[width=\GraphicsScalar]{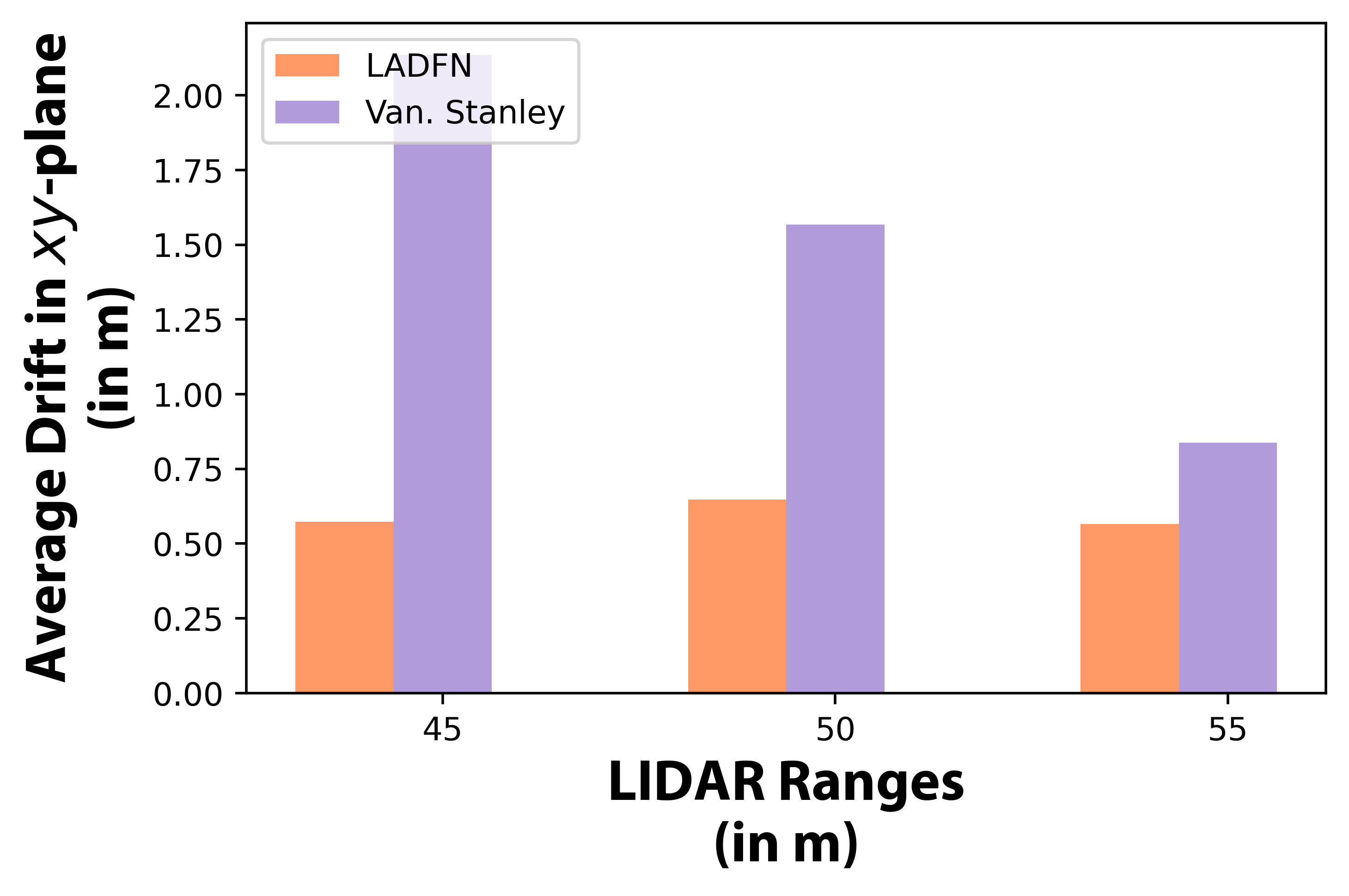}  
\end{subfigure}
\hfill
\begin{subfigure}[b]{\SubFigureScalar}
  \centering
  \includegraphics[width=\GraphicsScalar]{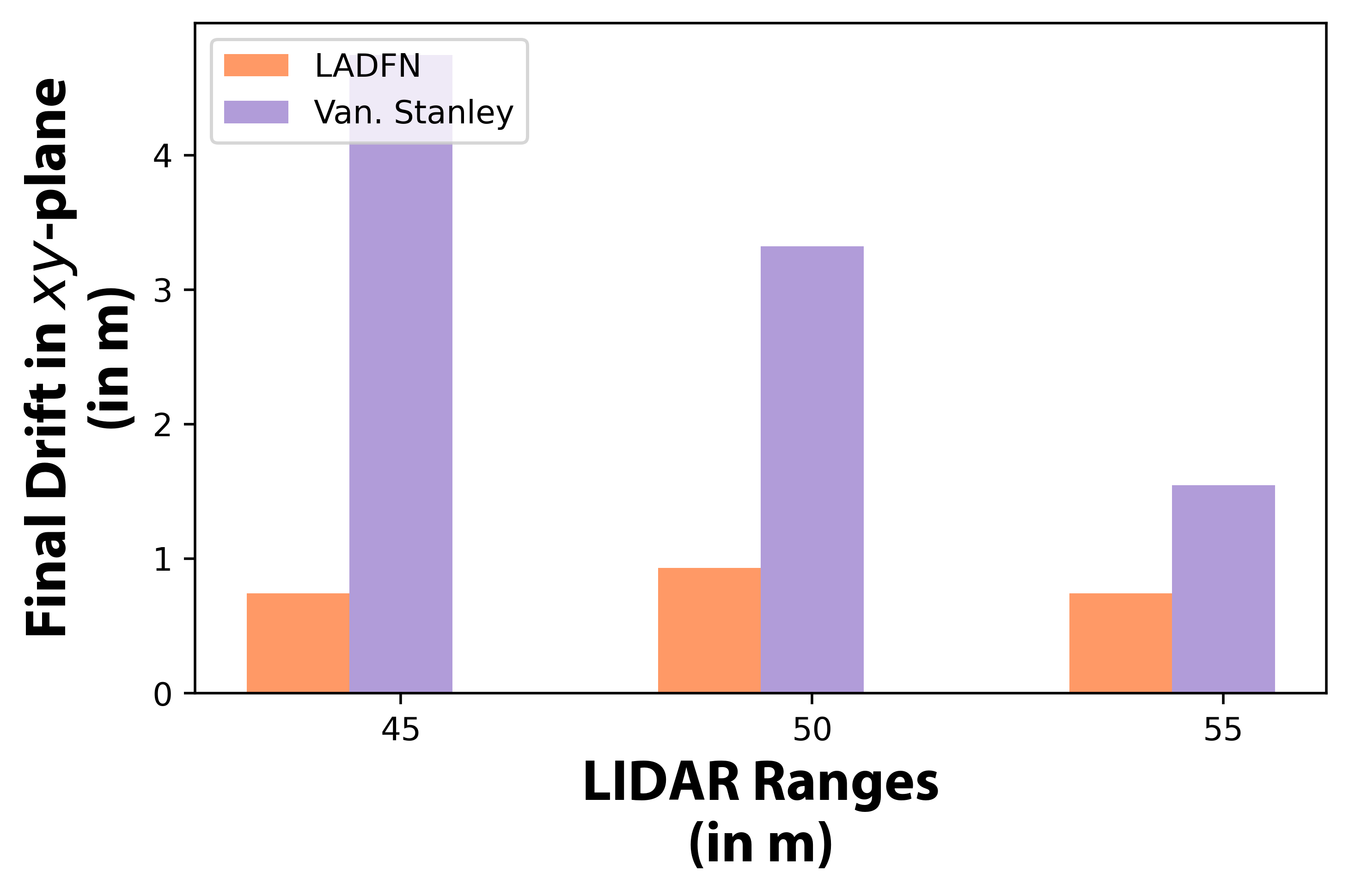}  
\end{subfigure}
\hfill
\begin{subfigure}[b]{\SubFigureScalar}
  \centering
  \includegraphics[width=\GraphicsScalar]{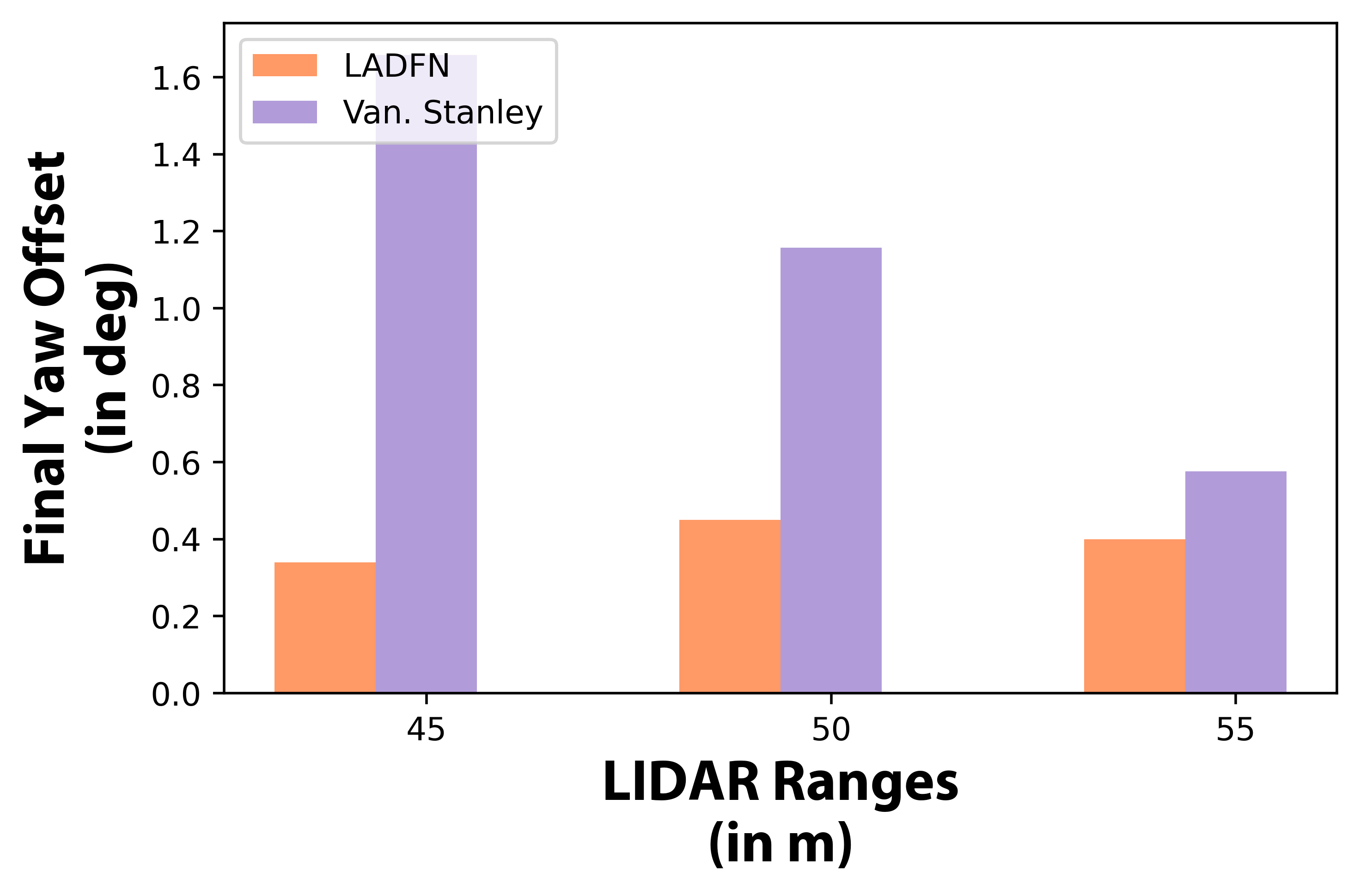}  
\end{subfigure}
\newline
\begin{subfigure}[b]{\SubFigureScalar}
  \centering
  \includegraphics[width=\GraphicsScalar]{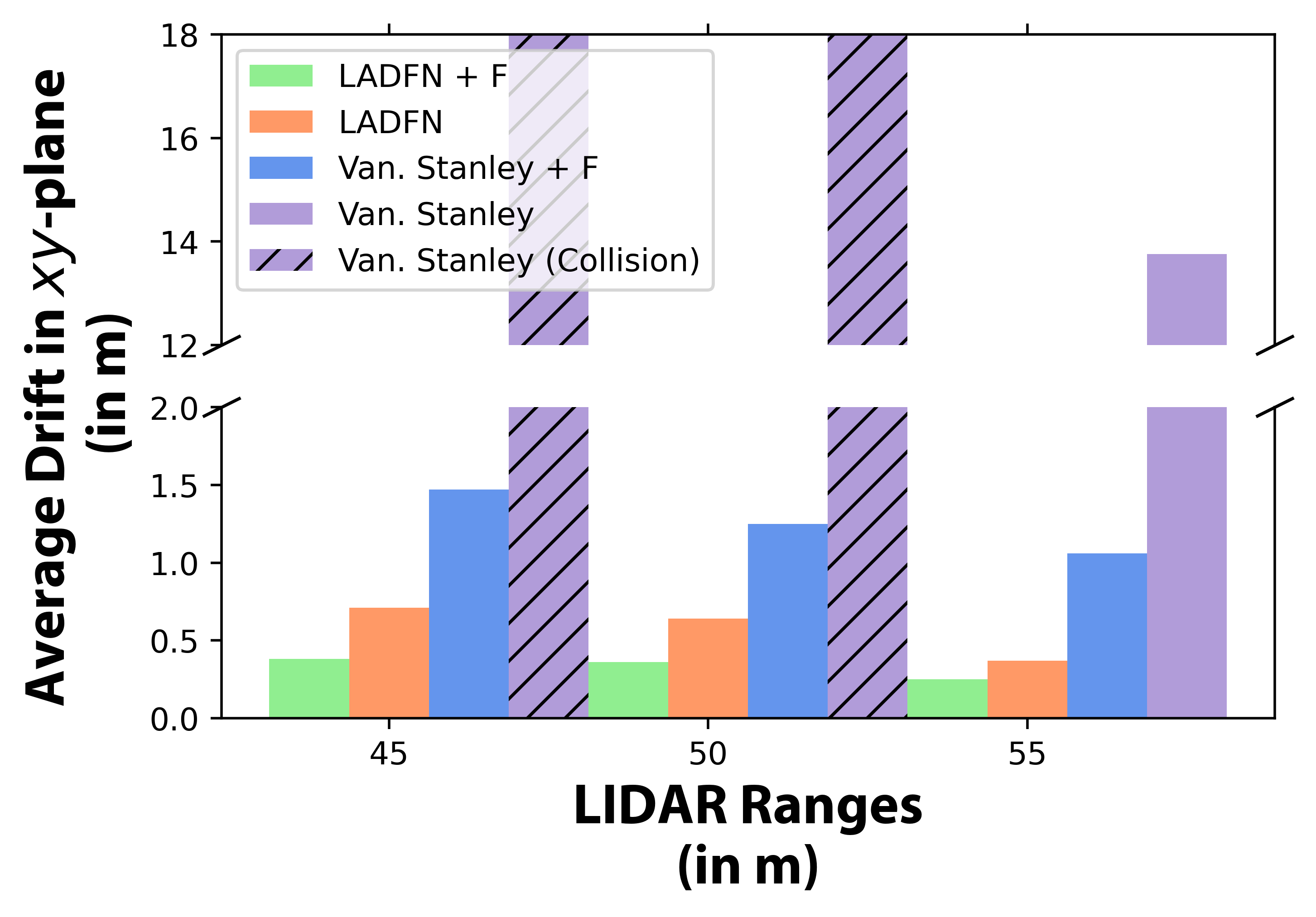}  
\end{subfigure}
\hfill
\begin{subfigure}[b]{\SubFigureScalar}
  \centering
  \includegraphics[width=\GraphicsScalar]{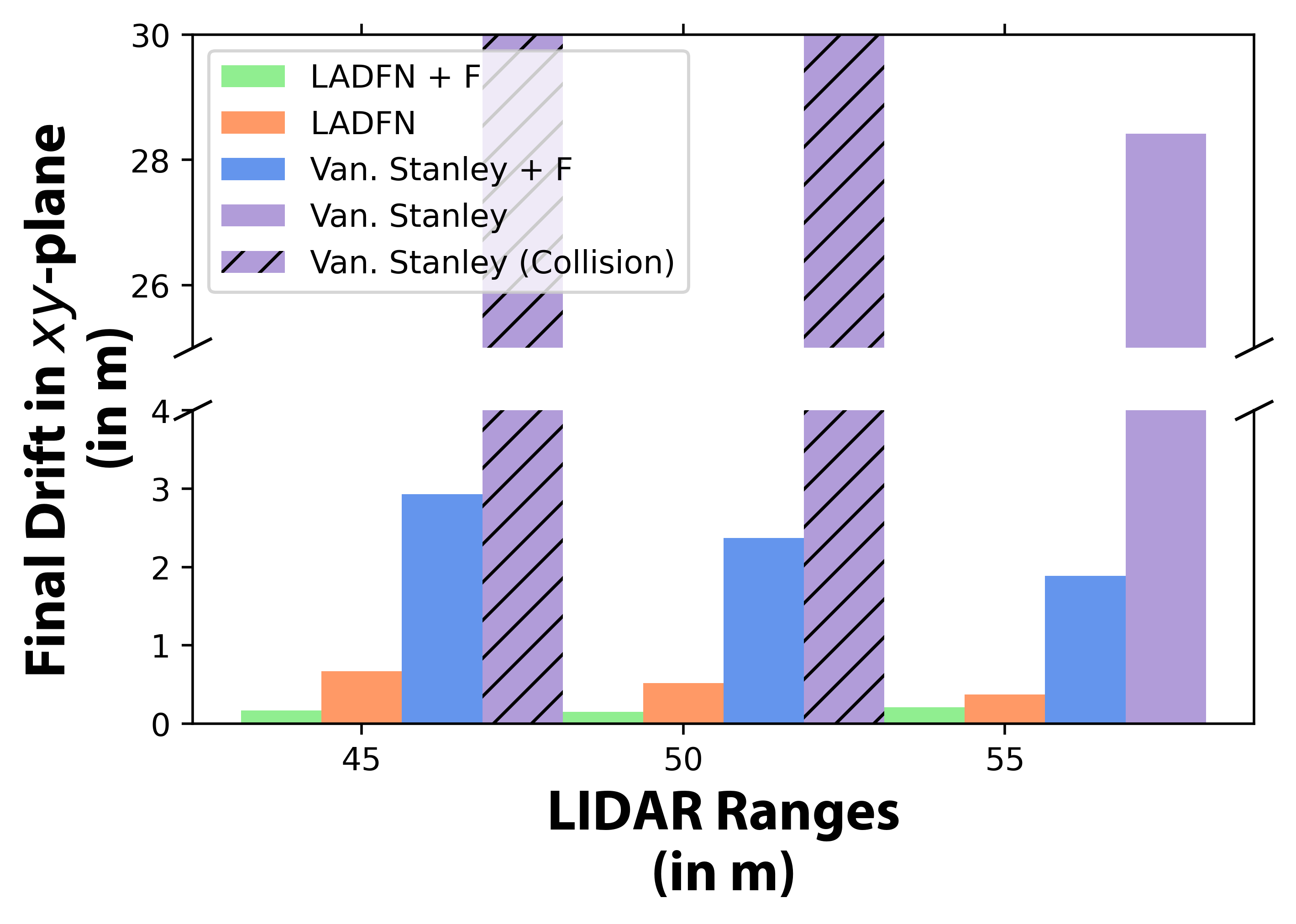}  
\end{subfigure}
\hfill
\begin{subfigure}[b]{\SubFigureScalar}
  \centering
  \includegraphics[width=\GraphicsScalar]{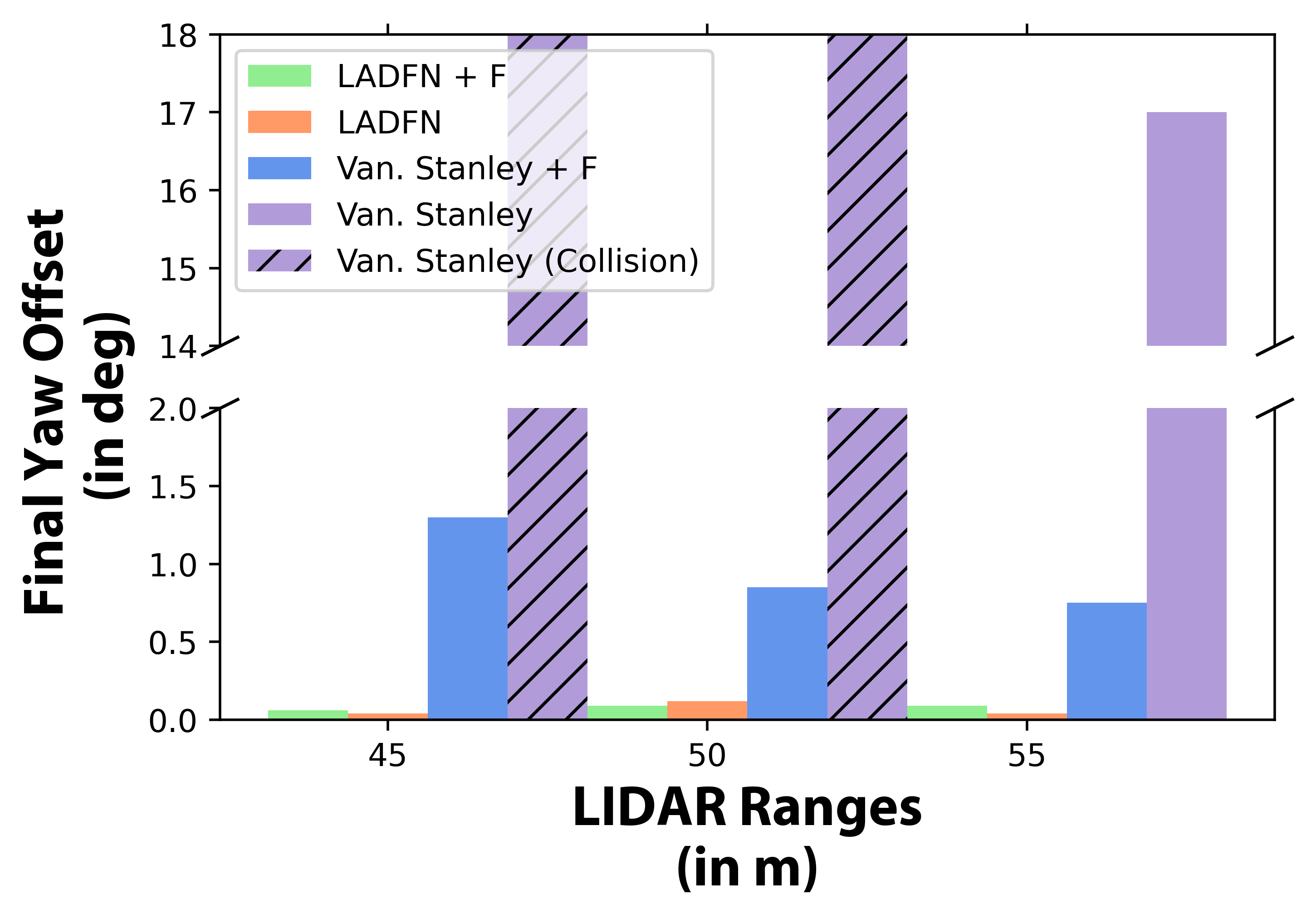}  
\end{subfigure}
\newline
\begin{subfigure}[b]{\SubFigureScalar}
  \centering
  \includegraphics[width=\GraphicsScalar]{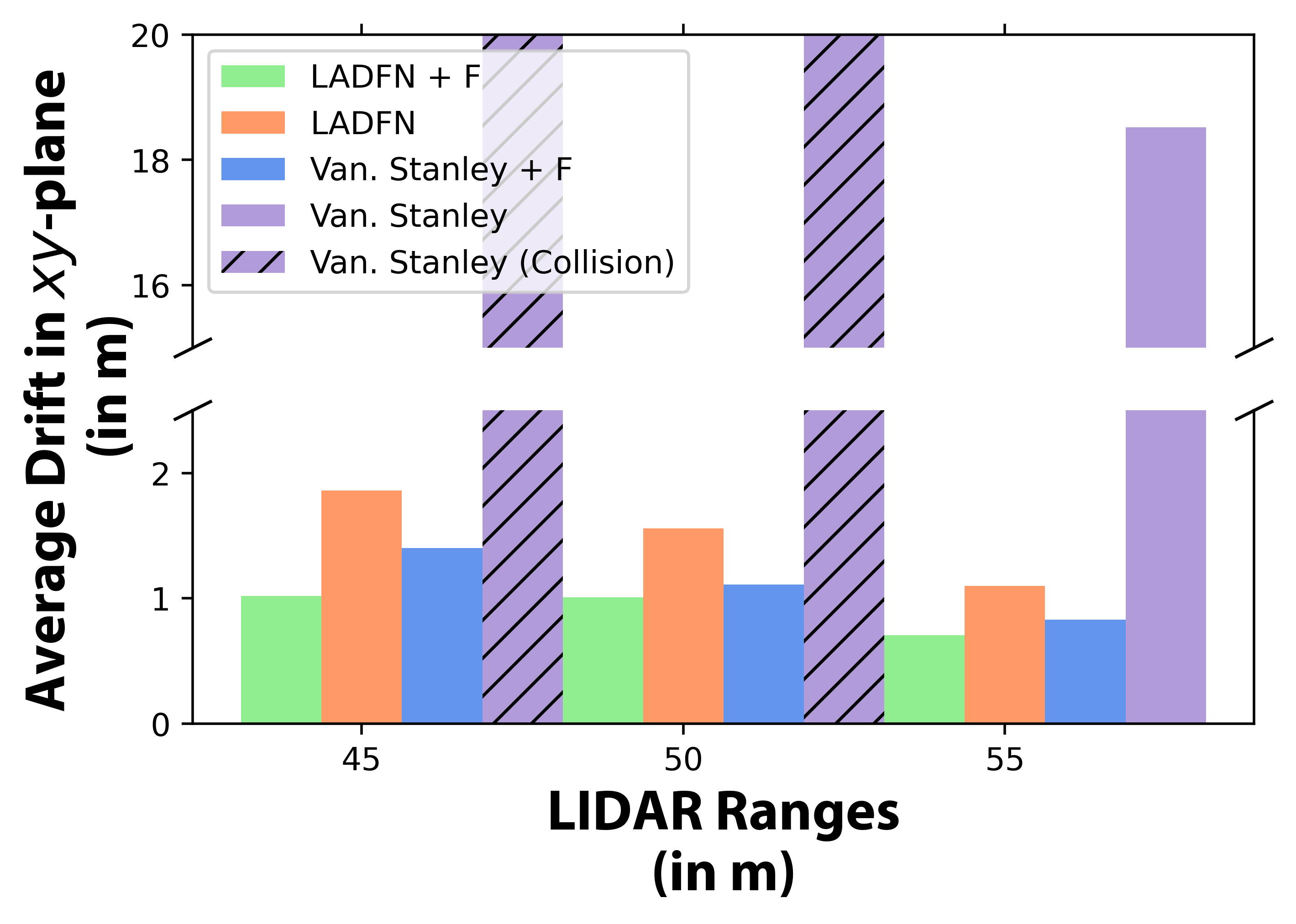}  
\end{subfigure}
\hfill
\begin{subfigure}[b]{\SubFigureScalar}
  \centering
  \includegraphics[width=\GraphicsScalar]{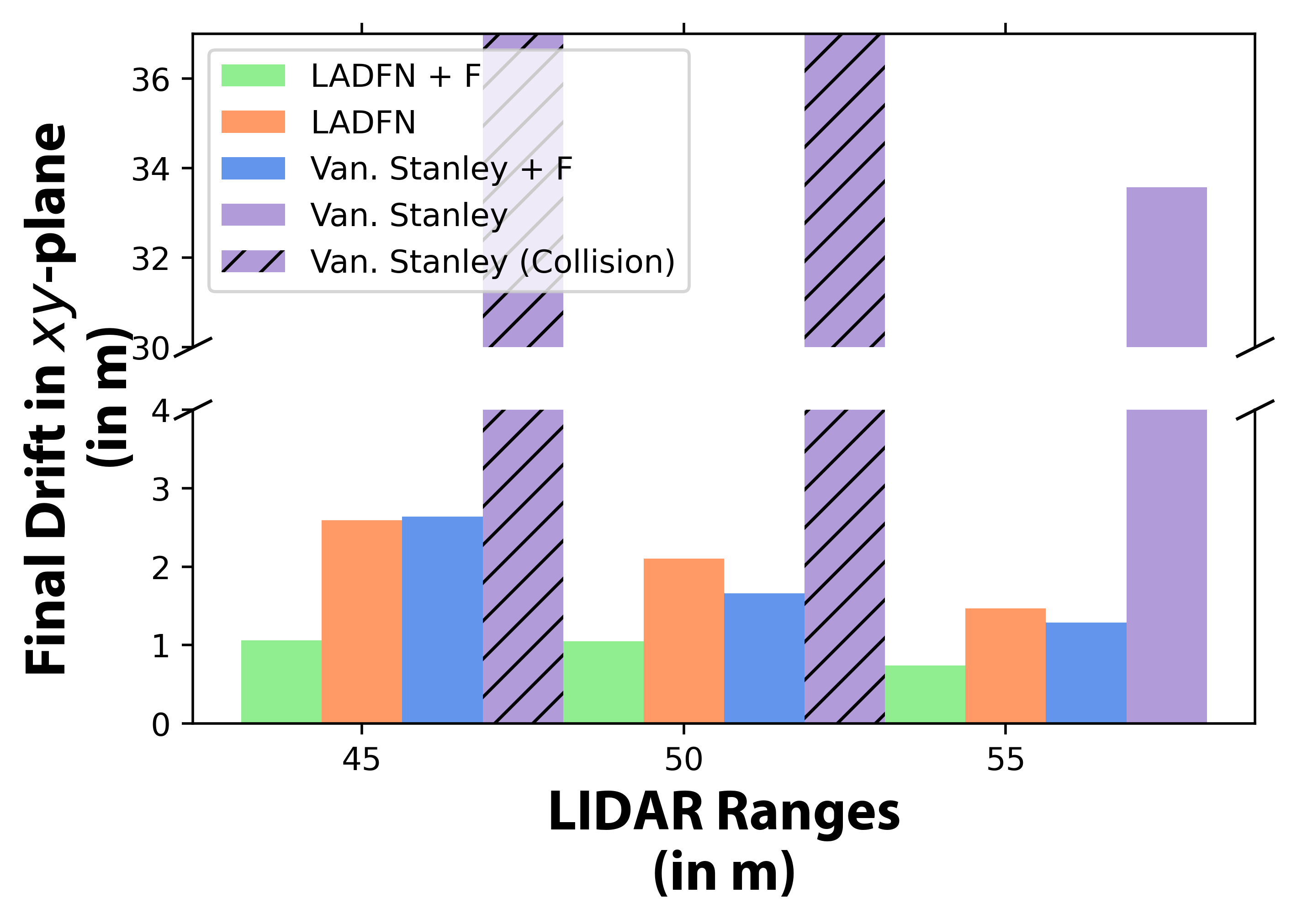}  
\end{subfigure}
\hfill
\begin{subfigure}[b]{\SubFigureScalar}
  \centering
  \includegraphics[width=\GraphicsScalar]{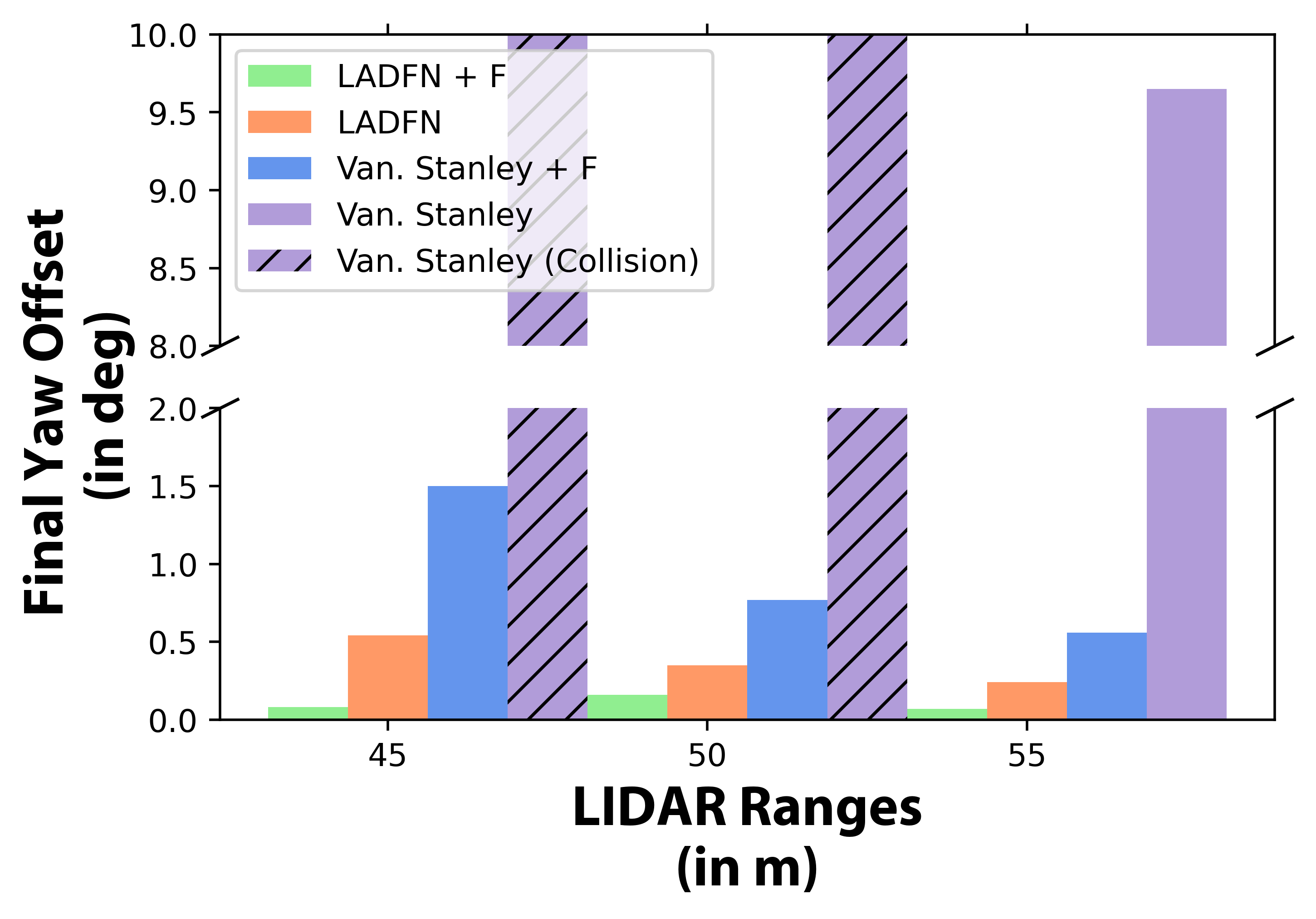}  
\end{subfigure}
\newline
\begin{subfigure}[b]{\SubFigureScalar}
  \centering
  \includegraphics[width=\GraphicsScalar]{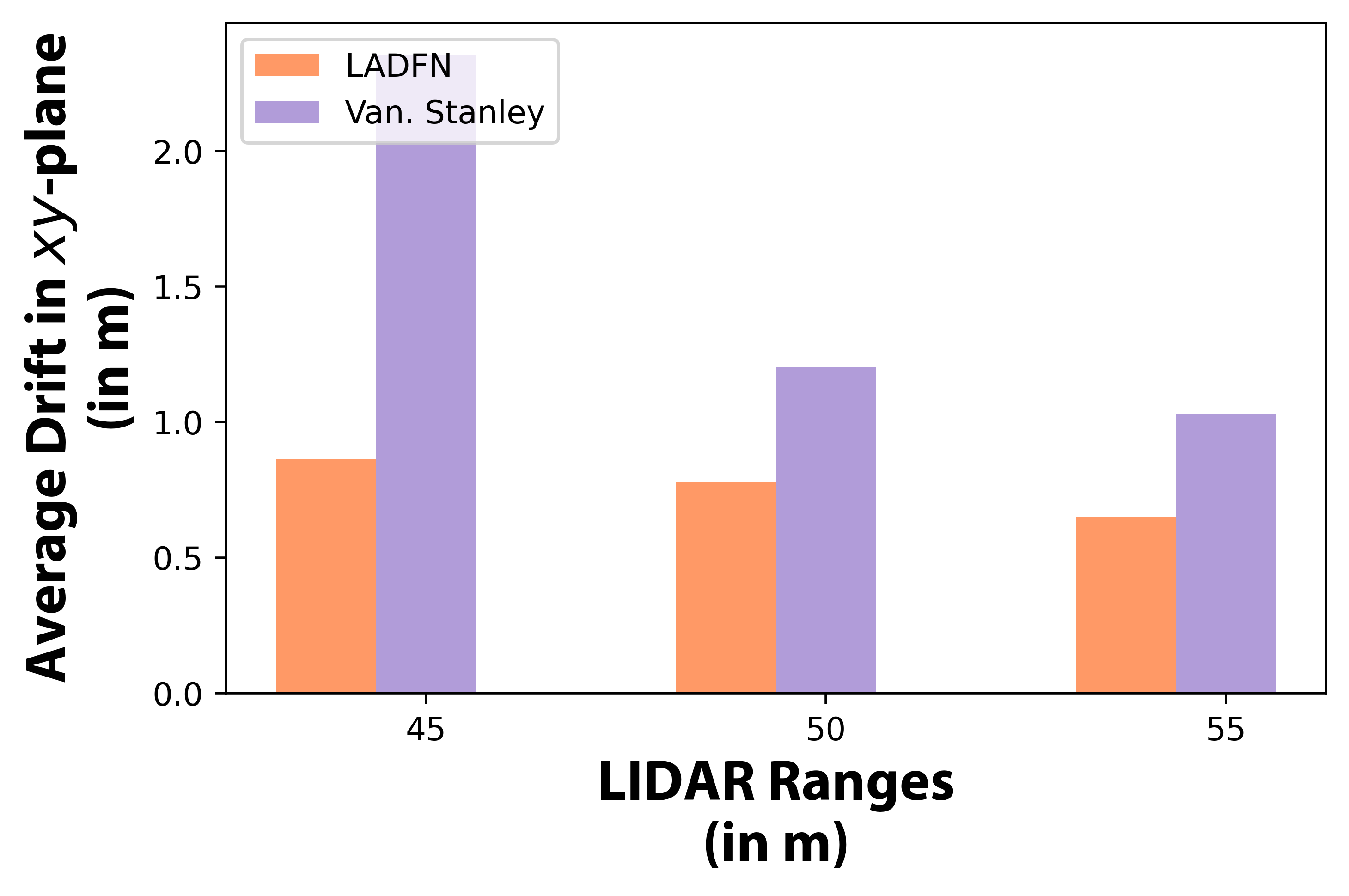}  
\end{subfigure}
\hfill
\begin{subfigure}[b]{\SubFigureScalar}
  \centering
  \includegraphics[width=\GraphicsScalar]{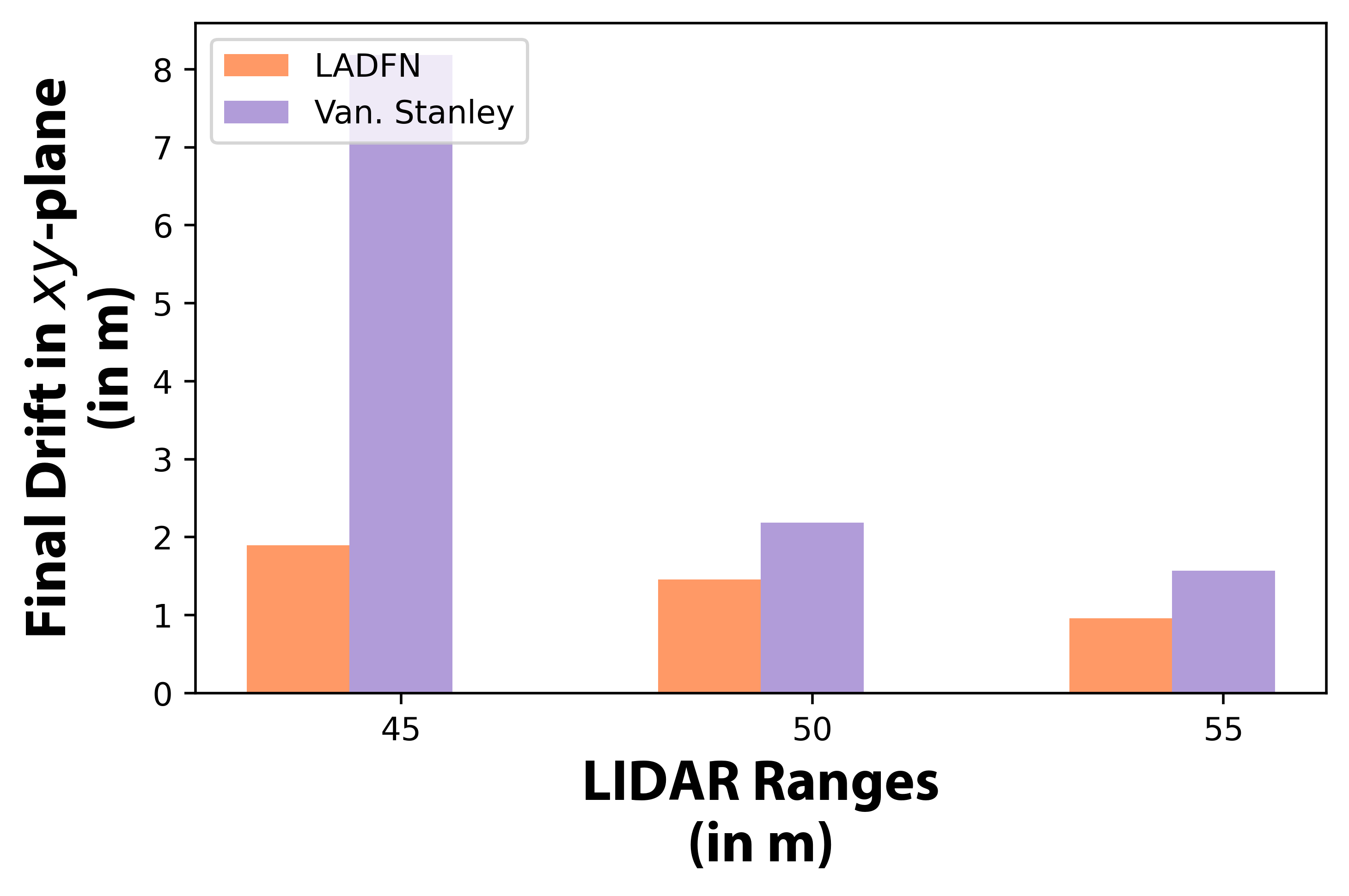}  
\end{subfigure}
\hfill
\begin{subfigure}[b]{\SubFigureScalar}
  \centering
  \includegraphics[width=\GraphicsScalar]{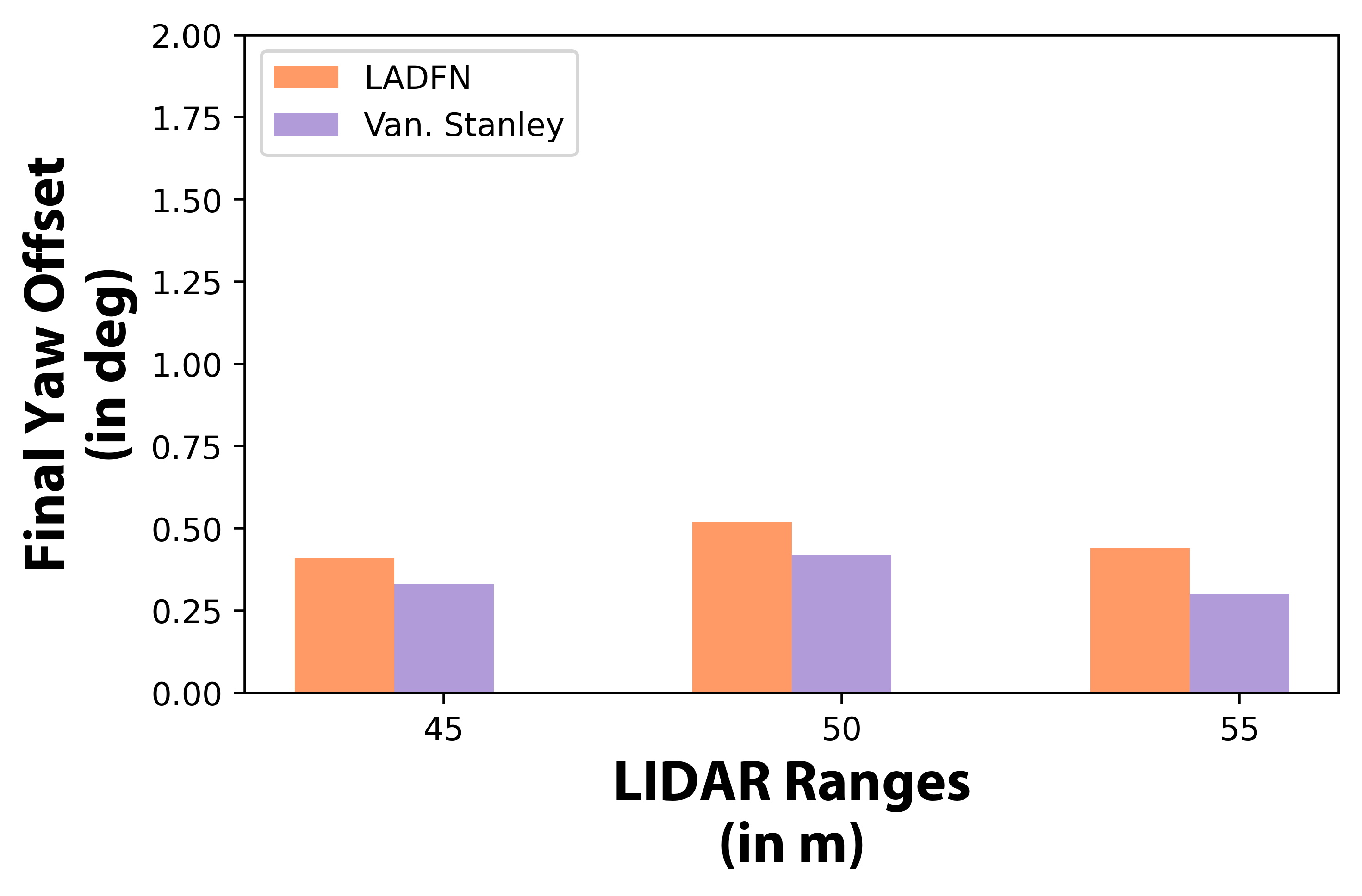}  
\end{subfigure}
\newline
\begin{subfigure}[b]{\SubFigureScalar}
  \centering
  \includegraphics[width=\GraphicsScalar]{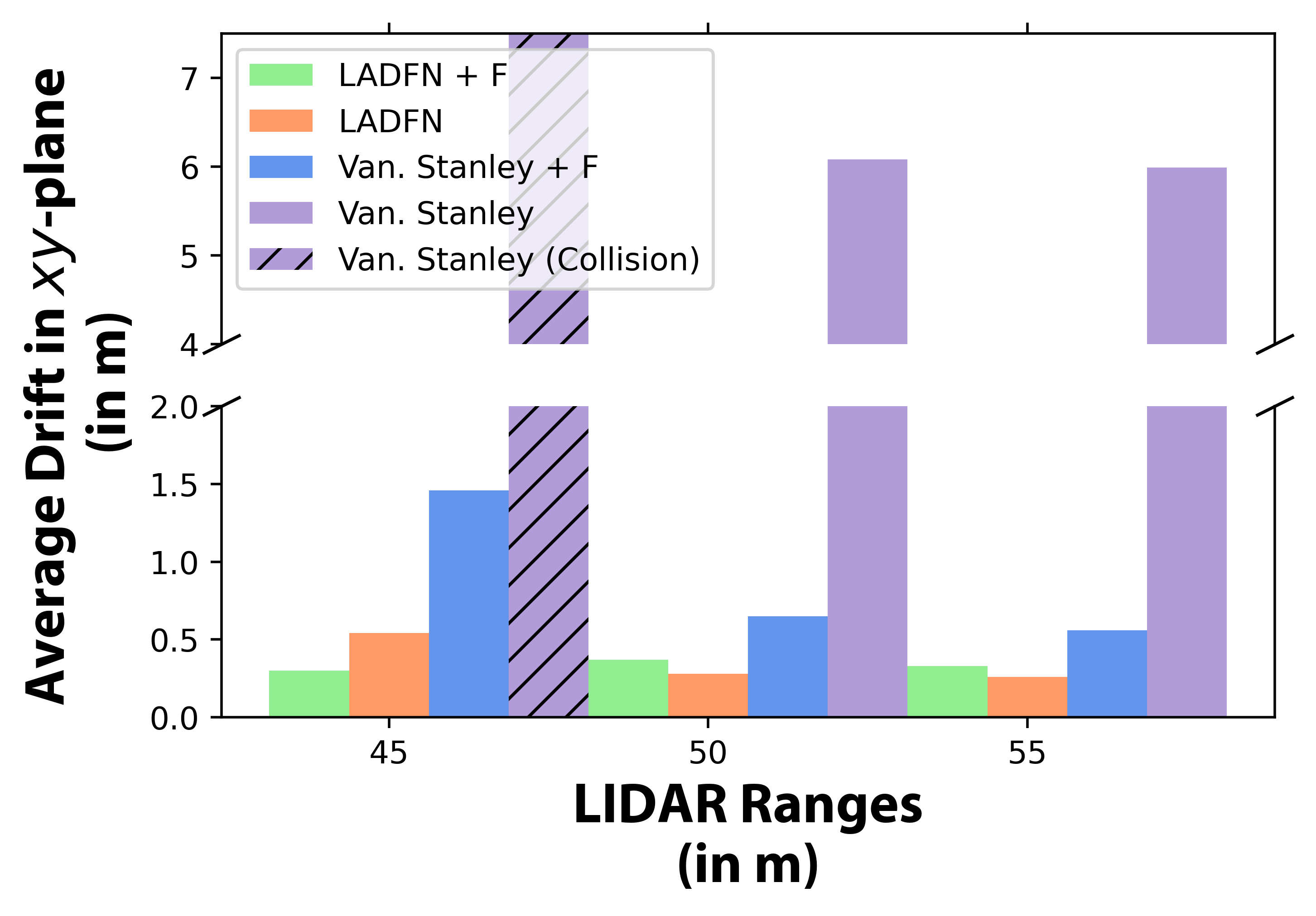}  
\end{subfigure}
\hfill
\begin{subfigure}[b]{\SubFigureScalar}
  \centering
  \includegraphics[width=\GraphicsScalar]{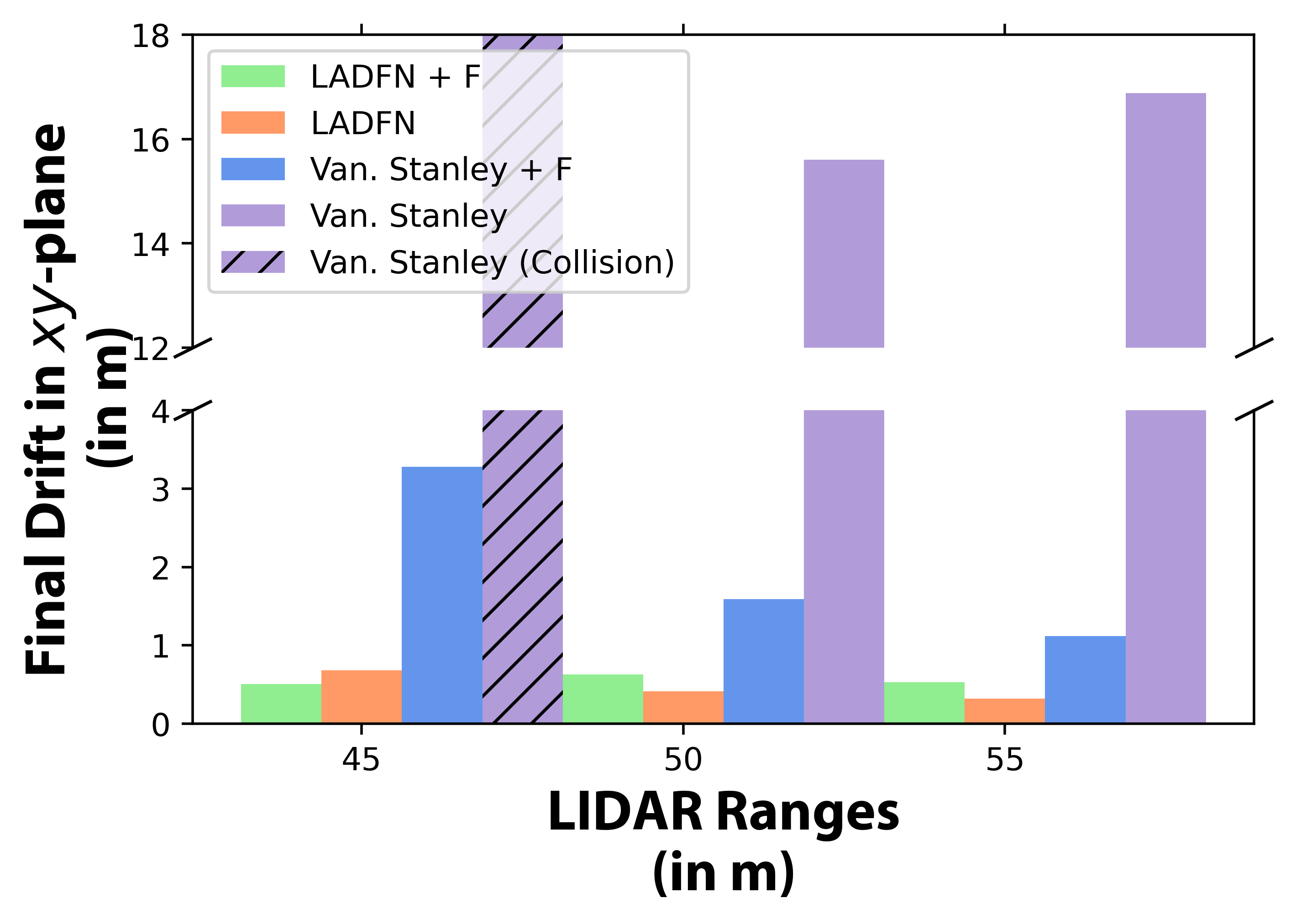}  
\end{subfigure}
\hfill
\begin{subfigure}[b]{\SubFigureScalar}
  \centering
  \includegraphics[width=\GraphicsScalar]{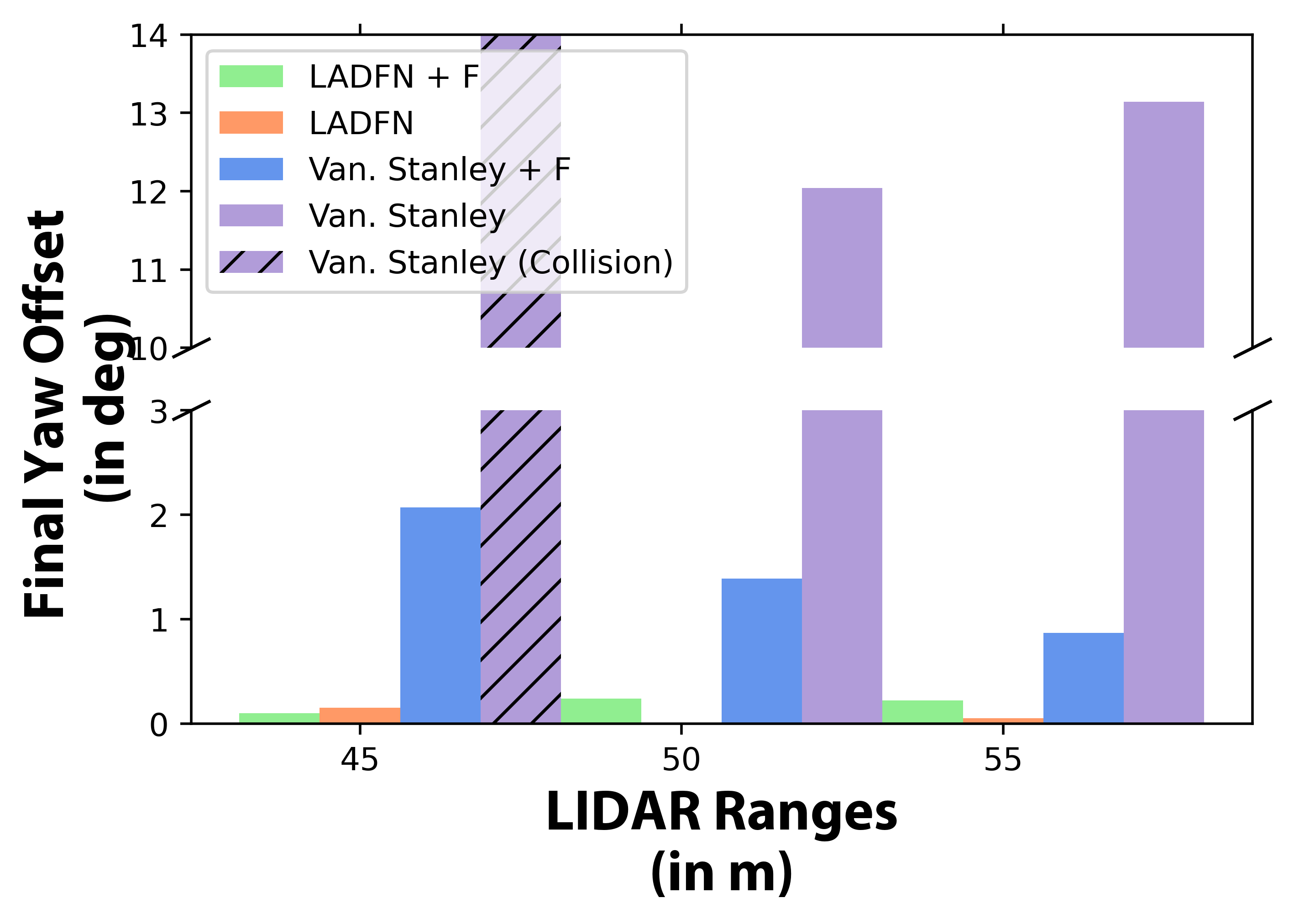}
\end{subfigure}
\setlength{\belowcaptionskip}{-15pt}
\caption{\figQuanResultsBarsOnlyCaption}
\label{fig:QuanResultsBarsOnly}
\end{figure*}
}

\usepackage{authblk}
\usepackage{enumerate}
\usepackage{hyperref}
\hypersetup{
    colorlinks=true,
    linkcolor=black,
    citecolor=black,
    urlcolor=blue
}

\title{\LARGE \bf
LADFN: Learning Actions for Drift-Free Navigation in Highly Dynamic Scenes
}
\author{Mohd Omama$^{1}$, Sundar Sripada V. S.$^{1}$, Sandeep Chinchali$^{2}$, K. Madhava Krishna$^{1}$%
\thanks{$^{1}$Robotics Research Center, IIIT-Hyderabad, India}%
\thanks{$^{2}$ECE Department, The University of Texas at Austin}%
\thanks{The authors thank  MathWorks India (Hyderabad) for their generous financial support.}%

}

\begin{document}
\maketitle
\begin{abstract}
We embark on a hitherto unreported problem of an autonomous robot (self-driving car) navigating in dynamic scenes in a manner that reduces its localization error and eventual cumulative drift or Absolute Trajectory Error, which is pronounced in such dynamic scenes. With the hugely popular Velodyne-16 3D \nameLIDAR\ as the main sensing modality, and the accurate \nameLIDAR-based Localization and Mapping algorithm, LOAM, as the state estimation framework, we show that in the absence of a navigation policy, drift rapidly accumulates in the presence of moving objects. To overcome this, we learn actions that lead to drift-minimized navigation through a suitable set of reward and penalty functions. We use Proximal Policy Optimization, a class of Deep Reinforcement Learning methods, to learn the actions that result in drift-minimized trajectories. We show by extensive comparisons on a variety of synthetic, yet photo-realistic scenes made available through the \nameCARLA\ Simulator the superior performance of the proposed framework vis-à-vis methods that do not adopt such policies.
\end{abstract}


\section{Introduction}
With rapidly increasing popularity of self-driving vehicles, there is a growing need for their adroit, drift-aware maneuvering in highly dynamic scenes, as drift particularly takes a hit in the presence of several moving objects. Highly dynamic scenes pose specific challenges as the signal is buried in noise for any kind of sensor based-localization/state estimation methods to prove effective. The autonomous vehicle needs to actively modify its behavior and navigation actions, often searching for spaces to move into from where it can reduce its drift and enhance its state estimation accuracy. To the best of our knowledge, there seems no literature that tackles this problem head on.
\figTeaser
\figPipeline

Therefore, in this paper, we present a novel active localization framework that seeks and learns actions that lead to and result in drift-minimized navigation in the presence of dynamic obstacles, as shown in Fig. \ref{fig:Teaser}. By suitably-crafted reward functions and PPO \cite{schulman2017proximal} as its Deep Reinforcement Learning architecture, action policies are learned that achieve this cause. We show by extensive comparisons in a variety of dynamic scenes the efficacy and superiority of this architecture in achieving drift-minimized navigation vis-à-vis methods that do not actively navigate to minimize state estimation errors. Henceforth, we call our system \textbf{LADFN} - \textbf{\underline{L}}earning \textbf{\underline{A}}ctions for \textbf{\underline{D}}rift-\textbf{\underline{F}}ree \textbf{\underline{N}}avigation. And for notational convenience, we refer to our autonomous vehicle as \textit{ego vehicle} or \textit{ego} interchangeably.

\subsection{Related Work}
State estimation or localization that involves computing the pose of the robot comes in various flavors with different sensing modalities such as those that estimate both the agent’s motion and the map \cite{dellaert2006square, kaess2008isam}, or those that only estimate the agent’s motion or odometry \cite{zhang2014loam, Kitt2010IV}. Prominent amongst these include the highly popular LOAM \cite{zhang2014loam} and its variants, with \nameLIDAR\ as the primary sensing modality or vision-based frameworks with or without direct availability of depth information \cite{labbe2019rtab, shan2018lego, Kitt2010IV, mur2015orb}. A modest portion of this literature is devoted towards handling dynamic actors. Herein, the popular theme is about filtering dynamic obstacles to enhance estimation accuracy irrespective of the sensing modality \cite{wang2014towards, kim2016effective, sun2017improving, bescos2018dynaslam, scona2018staticfusion, sun2018motion}. For example, authors in \cite{cheng2019robust} proposed to detect and eliminate the dynamic regions using a sparse motion removal (SMR) based on similarity and difference between frames that is fed to a Bayesian framework for motion detection and filtering. While \cite{kim2016effective} used background segmentation methods to filter dynamic agents, \cite{sun2017improving} used ego-motion compensation to detect dynamic regions in the current frame with subsequent vector quantization to segment moving regions in the scene. Other methods such as \cite{kundu2011realtime, reddy2015dynamic} jointly estimate stationary and moving agents in a multi-body framework. Amongst the popular \nameLIDAR\ based frameworks, \cite{pomerleau2014long} use long term mapping techniques to segment moving parts of the scene, while the stationary parts of the scene are used for localization and the moving agents are maintained in the map in the form of their velocity representations. In \cite{sun2016towards}, a distance-filtering approach is utilized to segment the dynamic contents of the scene while maintaining stationary parts as raw pointclouds.

The field of active localization, or actively navigating the robot or the ego to minimize localization errors, has received some attention in the literature. Active localization was popularized due to \cite{fox1998active, roy1999coastal} in their own ways. While \cite{fox1998active} focused on actively navigating towards less-aliased regions in the scene so as to converge towards a unimodal estimate of the state from a multi-modal belief due to perceptual aliasing, the main theme in \cite{roy1999coastal} was actively navigating to reduce the uncertainty of the unimodal state distribution. The complexity of active localization was analyzed in \cite{dudek1998localizing} and was shown to be NP-Hard, whereas a randomized algorithm for the same was presented in \cite{rao2005minimum}. The first and possibly only such formulation for active localization in a multi-agent setting was presented in \cite{bhuvanagiri2010motion}. However, none of the active frameworks talk of navigating an agent to reduce drift in the presence of moving agents.

This work distinguishes itself from all the prior art in several ways. For one, it is the first such method to tackle the problem of active navigation in dynamic on-road scenes. In contrast to the state estimation approaches, the present formulation showcases the need to go beyond filtering frameworks by learning actions to maneuver around dynamic actors in a manner that reduces drift. Unlike previous methods for active navigation that only handled stationary scenes, the current formulation is specifically tailored towards highly dynamic on-road scenes. 

We use deep RL to learn drift-minimizing behaviours even though alternate meta-heuristic approaches like GA, BAS have been used in robot control \cite{khan2022human}. These approaches suffer from convergence issues with increased search space in complex problems. A gradient-based RL system generally outperforms gradient-less meta-heuristic methods.


\subsection{Contributions}
\begin{enumerate}
    \item \textbf{Formulating Active Navigation as an RL Problem:} We formalize the problem of active navigation in the LIDAR setting using RL. 
    \item \textbf{Drift-Minimizing Reward Function:} 
    Our novel reward function facilitates learning of complex actions that simultaneously optimize for drift and control costs.
    We discuss this in detail in \ref{sec:Reward}. 
    \item \textbf{Integrating Learned Actions with a Classical Controller:} We combine the learned actions from RL with a Stanley Controller \cite{stanleycontroller} for path tracking. 
\end{enumerate}

\section{Pipeline Flow} \label{sec:PipelineFlow}
The complete end-to-end architecture of \namePaper\ is shown in Fig.~\ref{fig:Pipeline}. Our implementation has the following modules:
\begin{enumerate}
    \item \textit{Perception and Dynamic Obstacle Filtering}: We use an out-of-the-box implementation of LOAM \cite{zhang2014loam}. Then, we filter \nameLIDAR\ points originating from traffic vehicles.
    \item \textit{Waypoint Generation}: We use an RL model to choose actions that provide discrete waypoints for the ego to traverse. The model is unrolled for $n$ steps to get a trajectory.
    \item \textit{Trajectory Interpolation}: We fit a cubic spline on the trajectory to smoothen it.
    \item \textit{Low-level Control}: We use a Stanley Controller \cite{stanleycontroller} to perform path tracking on the trajectory.
\end{enumerate}

\subsection{Perception and Dynamic Obstacle Filtering}
\nameLIDAR\ Odometry and Mapping (\nameLOAM) \cite{zhang2014loam} is a SLAM technique \cite{durrant2006simultaneous} developed for 3D \nameLIDAR\ sensors. \nameLOAM\ splits the complex problem of optimizing several thousand variables simultaneously by splitting the mapping and localizing procedures into two: \textit{laser odometry} and \textit{laser mapping}. \textit{Laser odometry} runs at a high frequency and seeks to localize the \nameLIDAR\ by calculating its velocity in every frame. \textit{Laser mapping} runs at a lower frequency and finds edge-edge correspondences and plane-plane correspondences. These two processes are followed by iterative scan matching to get the translation $T$ and rotation $R$. \nameLOAM\ is a state-of-the-art odometry solution on several benchmark datasets, especially KITTI \cite{Geiger2012CVPR}. In our method, \nameLOAM\ runs in the background interacting with the \nameLIDAR\ points originating from \nameCARLA. We use \nameCARLA's \cite{Dosovitskiy17} provided \nameLIDAR\  segmentation to filter out points coming from dynamic vehicles and test our approach both in the presence and absence of this filtering. The odometry output from \nameLOAM\ is piped into the waypoint generation module, discussed in detail in Section \ref{sec:waypointgen}.

\subsection{Waypoint Generation} \label{sec:waypointgen}
The Waypoint Generation problem will be formulated as a Markov Decision Process (MDP) in Section \ref{sec:RLMethod}. While there are a plethora of deep RL algorithms to approximately solve this MDP, we chose PPO \cite{schulman2017proximal} since it is well-established and well-suited for discrete action spaces that our setting requires. We also note that the primary novelty of our paper is the MDP formulation for active navigation to reduce drift, and other RL algorithms should perform well in practice. For both the actor and the critic, we have a fully connected network with 6 hidden layers. The number of neurons in each hidden layer are: 64, 64, 32, 32, 16 and 16 respectively. We use ReLU activation along with a batch-size of 128 and a discount factor $\gamma$ of 0.9999. Our architecture conforms to the implementation of PPO in Stable-Baselines 3 \cite{stable-baselines3}.

We synthetically create a wide range of initial states. These states span across multiple combinations of ego positions, traffic positions, traffic speeds, positions of centroid of edge features, etc., During training, we assume that the position of centroid of edge features remains constant for that episode. This hypothesis is valid because our training episodes last for only 10 steps. We created an environment that takes as input at time $t$ the state $s_t$, and the action $a_t$, and outputs at time $t+1$ the next state $s_{t+1}$, and reward $r_{t+1}$. This environment was used to train an RL agent without directly interacting with the \nameCARLA\ \cite{Dosovitskiy17} Simulator. The episode is terminated either at the completion of 10 steps, or when the ego breaches lane boundaries, or when the ego collides with the traffic. 

\subsection{Trajectory Interpolation}
The waypoints obtained from PPO are then passed into a cubic spline planner for interpolation in order to obtain a smooth trajectory that our ego can traverse. During inference, we unroll the waypoint generation module $n$ times to get an $n$-step trajectory. We then fit a cubic spline on it using an existing framework provided by Sakai \textit{et al.} \cite{sakai2018pythonrobotics}.

\subsection{Low-level Control}
The final step in our process is to traverse the smooth spline and ensure that the ego is tracking the path with minimal drift. For this purpose we use a standard Stanley Controller \cite{stanleycontroller}, but it should be noted that our approach is invariant to the choice of the path tracking algorithm. The implementation of Stanley used is also provided by \cite{sakai2018pythonrobotics}.

\section{Model-Free RL for Drift Minimization} \label{sec:RLMethod}
Reinforcement Learning (RL) has been extensively used to solve a vast variety of Markov Decision Processes (MDPs). A canonical MDP comprises of a state space $\mathcal{S}$, an action space $\mathcal{A}$, a transitional probability, $\mathcal{P}$: $\mathcal{S}\times\mathcal{A}\times\mathcal{S}\rightarrow[0,1]$, that governs the transition from state $s_t$ to state $s_{t+1}$, a reward function, $\mathcal{R}$: $\mathcal{S}\times\mathcal{A}\times\mathcal{S}\rightarrow \mathbb{R}$, which describes a scalar reward that is associated with the transition from state $s_t$ to $s_{t+1}$ by taking an action $a_t$, and a discount factor $\gamma$ which controls the importance we give to future rewards.
When we solve the MDP, we find a policy $\pi$: $\mathcal{S}\rightarrow\mathcal{A}$, such that it maximises the value function $V_\pi$:
\eqnRLSol

Now, we will formulate \namePaper's waypoint generation problem as an MDP and provide a detailed account of the state space $\mathcal{S}$, the action space $\mathcal{A}$ and the reward $\mathcal{R}$. The dynamics of our MDP depend on the motion of the ego-vehicle as well as the uncontrolled motion of other traffic. Specifically, we cannot analytically model where other traffic will move, nor what the ego will observe at future time-steps from its \nameLIDAR\ sensors. Since we do not have an analytical model for the full dynamics of the MDP, we resort to model-free deep RL to solve for a control policy.
We first define a frame whose $X$-axis points along the direction of the road, and whose $Y$-axis is perpendicular to the direction of the road. The ego's starting coordinate in this frame will be of the form (0,  $y_e$). We refer to this frame as the \textit{RL frame} (Fig.~\ref{fig:rlframe}). 
\figRLFrame

\subsection{State}
We define the state of the RL agent as a vector such that it captures all necessary task-relevant information. The entries of the state space vector will be in \textit{RL frame} unless stated otherwise. The state vector is as follows:
\begin{itemize}
    \item $x_e$: x-coordinate of ego vehicle
    \item $y_e$: y-coordinate of ego vehicle
    \item $y_c$: y-coordinate of centroid of the edge features which is clipped and scaled to $[-1,1]$; edge features are discussed in Section \ref{reward_motivation}
    \item $edge_l$: left limit of the road
    \item $edge_r$: right limit of the road
    \item $tr_{p_i}$: Boolean representing the presence of $i^{th}$ traffic vehicle
    \item $x_{tr_i}$: x-coordinate of the $i^{th}$ traffic vehicle
    \item $y_{tr_i}$: y-coordinate of the $i^{th}$ traffic vehicle
    \item $v_{tr_i}$: velocity of the $i^{th}$ traffic vehicle.
\end{itemize}

\subsection{Action Space}
The RL agent has a set of discrete actions $\mathcal{A}$ that it can choose from. Each action $a\in\mathcal{A}$, represents the waypoint at the next time step for the vehicle in the vehicle’s frame of reference. The action $a$ is defined as a tuple $(a_x,a_y)$ where $a_x\in\{1, 2, 3, 5\}$ represents $x$-coordinate of the next waypoint, and $a_y\in \{-2,, -1, 0, 1, 2\}$ represents $y$-coordinate of the next waypoint in the vehicle's frame. We ensured that our action space reflects a spectrum of ways in which the vehicle could move: from slow-straight motion to fast-turning motion. 



\subsection{Motivation For Reward Shaping} \label{reward_motivation}
LOAM \cite{zhang2014loam} geometrically categorizes the \nameLIDAR\ points as edge or plane features which are then used in scan-matching. Edge features represent texture-rich regions which help in accurate scan-matching. We will refer to such regions as \textit{feature-rich.}
In order to sustain \nameLOAM's mapping process, and more importantly, reduce the absolute pose error (APE) in ego-motion, the first consideration is the ego's proximity to feature-rich regions. This can be argued as follows.

Consider two vehicles $ego_1$ and $ego_2$ and a feature-rich region $T_R$ as shown in Fig.~\ref{fig:featprox}, where $ego_1$ is closer to $T_R$ than $ego_2$. We draw tangents from the two vehicles to $T_R$. Let the angles that these two sets of tangents subtend with $T_R$ be $\theta_1$ and $\theta_2$ respectively.  Also, consider a planar \nameLIDAR\ which has a $360^\circ$ field view of a single plane and an angular resolution of $n$.  The number of \nameLIDAR\ rays per unit angle will then be $\frac{n}{360}$. Now, the number of \nameLIDAR\ rays that interact with the feature-rich region in both the cases will be $\frac{n\times\theta_1}{360}$ and $\frac{n\times\theta_2}{360}$ respectively. Since, $ego_1$ is closer to $T_R$ than $ego_2$, $\theta_1$ will be larger than $\theta_2$, indicating that more \nameLIDAR\ rays will interact with $T_R$ in the first case leading to its more dense reconstruction. The same argument can be extended for 3D \nameLIDAR\, which is the primary hardware setting in our proposed solution.  Therefore, when the ego is in the proximity of texture-rich regions, \nameLOAM\ can more densely reconstruct them since the incoming pointcloud to \nameLOAM\ will have more edge features contained in it.

\figFeatureProx

The second aspect that plays an impact on the APE is the distance from traffic vehicles. \nameLIDAR\ points from dynamic objects lead to erroneous registrations in the scan-matching algorithm; in our case, the dynamic objects are said traffic vehicles. Further, proximity to dynamic vehicles can also occlude visibility of other texture-rich regions. Follow-up papers \cite{thomas2019delio, zhao2019robust} on \nameLOAM\ have argued the same. Further, Thomas \textit{et al.} \cite{thomas2019delio}, have specifically shown that the proximity to dynamic vehicles leads to substantial error in the rotational measurement, $R$. Our experimentation in dynamic scenes also resonates with the findings of these works but unlike us, none of them address drift-minimizing behaviors. 

To handle the problems described above, the ego should be aware of the feature-rich regions and plan its trajectory accordingly. It should also avoid proximity with traffic wherever possible. 
Our proposed solution comprises of an RL agent that generates waypoints which are then traversed with the aid of a low-level controller. The input to the RL agent is a state that includes the centroid of edge features, the ego's pose, and the traffic vehicles' poses. We calculate the centroid using queried edge features obtained through \nameLOAM. The pose of the ego is directly queried from \nameLOAM. We also assume that we have the speed and pose information of the traffic readily available due to recent advancements in \nameLIDAR-based semantic segmentation and tracking \cite{milioto2019rangenet++}.

\subsection{Reward} \label{sec:Reward}
The reward was designed with two primary objectives: navigation and perception. The navigation objectives comprise of the follows.
\begin{enumerate}
    \item Forward motion: We incentivize the ego to move forward towards the goal.
    \item Road boundary constraint: $edge_l \leq y_e \leq edge_r$, i.e., the ego should move within the road boundaries.
    \item Collision constraint: $\lVert (x_{tr_i}, y_{tr_i}) - (x_e, y_e) \rVert \geq 3$ meters, i.e., the ego must not collide with the traffic.
\end{enumerate}

The perception objectives comprise of the follows.
\begin{enumerate}
    \item Feature proximity: \textit{minimize} $\vert y_e - y_c \vert$, i.e., the ego should try to maintain proximity with the feature-rich regions.
    \item Traffic distance: $\lVert (x_{tr_i}, y_{tr_i}) - (x_e, y_e) \rVert \leq 10m$, where the ego should try to avoid proximity with traffic. 
\end{enumerate}

Our reward $\mathcal{R}$ models both the objectives described above and is defined as:
\eqnReward
where $\mathcal{P} = \lor {p}_i$, which is the logical OR over all the \textit{traffic-proximity} Boolean $p_i$. $p_i$ will be described in detail in the \textit{Traffic Reward} section. We now explain the other three terms in equation~\eqref{eq:TotalReward}.
\subsubsection{The Goal Reaching Reward, $\mathcal{G}$} $\mathcal{G}$ returns a positive reward for forward motion along the $X$-axis towards the goal. Although a simple term, this ensures that the ego prioritizes moving towards the finish line besides pursuing feature-rich regions and avoiding traffic. 
The other navigation constraints like traffic and road limits are also modeled here. The ego is penalized if it collides with the traffic or moves out of the road, and the episode terminates. 
\subsubsection{The Feature Reward, $\mathcal{F}$}
The feature reward needs to be shaped in a way that the ego is rewarded if it maintains proximity with feature-rich regions. To achieve this, we need two behaviors that are captured in two terms in \eqref{eq:FeatReward} as follows:

\paragraph{The first term} The ego should move towards the left if the left side of the road is feature-rich and vice-versa. The lateral position of the feature-rich region is described by $y_c$, which is clipped and scaled to a normalized range of $[-1,1]$. A value of $-1$ means the centroid of features is towards the left end of the road, and a value of $1$ means the centroid is towards the right. To formulate the feature reward, we also need $y_e$ in the range $[-1,1]$. The term $(2((y_e-{y_e}_\mathrm{min})/({y_e}_\mathrm{max} - {y_e}_\mathrm{min})) - 1)$ handles that. We first raise $y_c$ to an odd power, $O$, and then multiply it with normalized $y_e$. This means that the reward will be positive if both $y_c$ and normalized $y_e$ have the same sign, implying they are on the same side of the road; and the reward will be negative if they are on the opposite sides of the road. Further, $y_c^O$ forces the term following it to quickly drop to 0 when $y_c$ is close to 0, ensuring that this \textit{first term} has no impact when there are feature-rich regions on both sides of the ego.

\paragraph{The second term} If there are features on both sides of the road, the ego must prevent unnecessary lateral motion. To achieve this, we multiply the absolute of our lateral motion $|a_y|$ with $(1 - \vert y_c^O \vert)$. In this case, $ \vert y_c^O \vert$ will be close to zero since features are on both sides and $(1 - \vert y_c^O \vert)$ will tend to $1$. Further, $(1 - \vert y_c^O \vert)$ drops to zero when $y_c$ is close to 1, ensuring that this second term has no impact when there are features on only one side of the road (left or right).

The complete mathematical formulation of the feature reward is given by:
\eqnFeatureReward
where $K_1$ and $K_2$ are constants. 
\subsubsection{The Traffic Reward, $\mathcal{T}$}
We define a \textit{traffic-proximity} region $\lVert (x_{tr_i}, y_{tr_i}) - (x_e, y_e) \rVert \leq 10m$ where the effect of traffic is large on localization. The traffic reward should be formulated such that the ego tries to move out of this region. We design a reward which incentivizes the ego if the relative difference in forward direction between the ego and the traffic  increases. It also returns a positive reward if the ego increases its lateral distance with the traffic car, and a negative reward otherwise. The traffic reward can be mathematically written as:
\eqnTrafficReward
where $p_i$ is a Boolean which is 1 only when the ego is in a \textit{traffic-proximity} region, otherwise this traffic-vehicle has no impact on the reward, and $K_3$ and $K_4$ are constants. The summation term outside accumulates the rewards/penalties from each traffic vehicle $i$. 

The waypoint generation MDP is solved using deep RL using an architecture described in Section \ref{sec:waypointgen}. We now evaluate our RL-based active navigation framework, LADFN, in challenging autonomous driving scenarios. 




\section{Experiments} \label{sec:Experiments}

\subsection{Experimental Setup}
The performance of \nameLIDAR\ odometry algorithms can depend a lot on the physics of the environment as well as that of the ego. Hence, we use \nameCARLA\ simulator \cite{Dosovitskiy17} for our experimentation as it emulates real-world physics accurately. It is also a state-of-the-art high-fidelity simulation environment for autonomous driving and so it is ideal for our application. We simulate a 3D \nameLIDAR\ in \nameCARLA\ with 16 channels and a vertical field of view of $30^{\circ}$ ($-15^{\circ}$ to $15^{\circ}$). The \nameLIDAR\ emits 300,000 points per second which are evenly distributed over all channels. While running our experiments, we use three different \nameLIDAR\ ranges: 45m, 50m, and 55m. These ranges correspond to the the maximum distance that can be measured by the \nameLIDAR.\ The variety of \nameLIDAR\ ranges during the experiments helps us analyze the performance of our method for various sensor ranges.

\figQualResults

\subsection{Scene Description}
We have created a test bed of 5 different scenes in \nameCARLA\ \cite{Dosovitskiy17}. These scenes have a run length ranging from 100 to 500 meters, and include static as well as dynamic scenarios. We have created the scenes such that they have varying distributions of feature-rich regions and dynamic vehicles. Scenes 1 and 4 are static and don't have any dynamic traffic actors whereas scenes 2, 3 and 5 contain moving traffic cars or vans around the ego. The results shown have up to two dynamic vehicles, but the method is easily scalable to an arbitrary traffic count.
\figQuanResultsBarsOnly
\vspace*{-3pt}
\subsection{Metrics and Benchmarks}
We benchmark our approach against what we call the \textit{Vanilla Stanley Controller}. With \textit{Vanilla Stanley}, the ego will not actively change its direction or speed to account for drift minimization. The ego will drive with a speed that is similar to the traffic around it. Hence, it represents  \textit{perception-unaware control}, a realistic scenario where the ego is tracking a trajectory without having any idea of the perception goals. Since there is no previous work in this area, Vanilla Stanley is an ideal benchmark to compare against. 

In dynamic scenes, a general approach to decrease trajectory error is to filter out dynamic vehicles from the \nameLIDAR\ scans. Hence we have another benchmark which we refer to as  \textit{Vanilla Stanley + F} which is just a \textit{Vanilla Stanley Controller} running with filtering enabled. We test and compare our proposed approach both in the presence and absence of dynamic obstacle filtering. This helps us understand that even when filtering is available, active navigation is essential to further reduce drift in the trajectory.

We use three different metrics for evaluating our approach:
\begin{enumerate}
    \item \textit{Average Drift:} The average translation drift in $xy$-plane for the entire trajectory.
    \item \textit{Final Drift:} Final translation drift in $xy$-plane at the end of the trajectory.
    \item \textit{Rotational Offset:} Yaw offset with respect to the ground-truth yaw at the goal location.
\end{enumerate}
\textit{Metric 1} gives a general idea of the quality of localization achieved for the entire trajectory. \textit{Metric 2} gives an idea of how the methods will perform on goal-reaching tasks. \textit{Metric 3} serves two purposes: like \textit{Metric 2}, this is also a representation for goal-reaching tasks, and further, it gives an idea of the quality of the future trajectory, since rotational errors in odometry have a compounding effect on the future trajectory.

\subsection{Qualitative Results}
Fig.~\ref{fig:QualResults} shows a qualitative view of the trajectory and drift in all the five scenes. In the right column, as the estimated LOAM trajectory in red overlaps the ground-truth trajectory in black, unlike the baseline shown in the left column where the red deviates from the black, it can be clearly seen that with our approach, the drift is substantially reduced in every case. We can see that the ego tries to maintain proximity with the feature-rich regions shown in green. Rows 2, 3, and 5 correspond to dynamic scenes with filtering enabled and so traffic vehicles are not visible in the \nameLIDAR\ scans shown in the Fig.~\ref{fig:QualResults}. We observe that even in the presence of filtering, our approach outperforms the baseline, highlighting the need for active navigation in drift minimization. 

We achieve this due to the complex behaviours learned by our model, the examples of which can be seen in the following link: \url{https://mohdomama.github.io/LADFN/}. 

\subsection{Quantitative Results}


We depict the quantitative comparisons of our approach with the baselines in the five rows of Fig.~\ref{fig:QuanResultsBarsOnly} corresponding to the five scenes. Rows 1 and 4 do not have dynamic actors, and the comparison is between \textit{LADFN} and \textit{Vanilla Stanley}, shown with orange and violet bars. Rows 2, 3 and 5 correspond to the scenes populated with dynamic actors. Here, we include the filtering module in both the Stanley Controller and LADFN, denoted as \textit{LADFN + F} and \textit{Vanilla Stanley + F}, and shown with green and blue bars. In some scenes, the LOAM estimates have significantly deviated resulting in collision for the baseline \textit{Vanilla Stanley}, these are shown with hatched violet bars.

Our first step is to show that \namePaper\ leads to a lower average drift than benchmark competitors. This is clearly shown in column 1. In static scenes, we get up to $3.7\times$ improvement in average drift using \namePaper. In dynamic scenes, with filtering enabled, we get up to $4.8\times$ improvement in average drift using \namePaper, while with filtering disabled, the drift in \textit{Vanilla Stanley} is often so high that it leads to collision. This is not the case with \namePaper\ which remains relatively unaffected, even with filtering disabled. It should also be noted that the impact of dynamic vehicles and feature-proximity on benchmark competitors gets pronounced as the \nameLIDAR\ range decreases, while \namePaper\ is resilient to these changes in \nameLIDAR\ range.

The second column and the third column show the final drift and the final yaw offset at the end of the run respectively. They give an idea of the goal-reaching capabilities of all the approaches along with an intuition of how bad the future trajectory would be affected if the run was continued. We see that \namePaper\ outperforms the benchmarks here with similar patterns as in column 1. 

\section{Conclusion}
Future autonomous vehicles must actively navigate to reduce their localization and perception errors as well as optimize for conventional metrics such as obstacle avoidance and time-efficient navigation. This paper presents a principled method, based on deep RL, to synthesize navigation behaviors that significantly reduce localization error for autonomous vehicles using state-of-the-art \nameLIDAR\ SLAM methods. Our key technical insight was to adopt a hierarchical RL approach, which focuses on selecting drift-minimizing waypoints that can be efficiently tracked by a low-level Stanley Controller \cite{stanleycontroller}. 

In future work, we plan to implement our perception-aware RL navigation algorithm on a small mobile robot that operates in crowded, dynamic scenes. Moreover, we plan to work towards theoretical guarantees based on a model predictive control (MPC) framework where a robot must jointly optimize for safe, time-optimal waypoint control as well as minimizing localization errors. We intend to explore CEM-MPC based batch optimization over GPU \cite{aks_ral_behaviour_22} to manifest complex behaviors in real-time.  

\bibliographystyle{IEEEtran}
\bibliography{bibs}
\end{document}